\documentclass[10pt,twocolumn,letterpaper]{article}

\usepackage{iccv}              % To produce the CAMERA-READY version
\usepackage{colortbl}
\PassOptionsToPackage{table}{xcolor}
\usepackage{xcolor}
\definecolor{lightgreen}{RGB}{144, 238, 144}

\usepackage{graphicx}
\usepackage{multirow} 
\usepackage{makecell}
\usepackage{booktabs}
\usepackage{soul}
\usepackage{pifont}
\usepackage{hhline}
\usepackage{amsmath}
\usepackage{cases}
\usepackage{mathtools}
\usepackage{float} 
\usepackage[accsupp]{axessibility}  % Improves PDF readability for those with disabilities.
\newcommand{\bestimprovered}[2]{#1$_\text{\textcolor{red}{~(+#2)}}$}

\definecolor{iccvblue}{rgb}{0.21,0.49,0.74}
\usepackage[pagebackref,breaklinks,colorlinks,allcolors=iccvblue]{hyperref}

\def\paperID{10423} % *** Enter the Paper ID here
\def\confName{ICCV}
\def\confYear{2025}

\title{Plug-in Feedback Self-adaptive Attention in CLIP for \\ Training-free Open-Vocabulary Segmentation}

\author{Zhixiang Chi$^{1}$, ~~
Yanan Wu$^{2}$, ~~
Li Gu$^{3}$,~~
Huan Liu$^{4}$,~~
Ziqiang Wang$^{3}$,~~
Yang Zhang$^{5}$ \\
Yang Wang$^{3}$, ~~
Konstantinos N Plataniotis$^{1}$
\\
[0.1cm]
${^1}$University of Toronto \quad 
${^2}$ China Agricultural University \quad
${^3}$ Concordia University \quad \\
${^4}$ McMaster University \quad
${^5}$ Beijing Jiaotong University \\
{\tt zhixiang.chi@mail.utoronto.ca} \quad \href{https://github.com/chi-chi-zx/FSA}{\color[HTML]{ED028C}https://github.com/chi-chi-zx/FSA}\\
}

\begin{document}
\maketitle
\begin{abstract}
CLIP exhibits strong visual-textual alignment but struggle with open-vocabulary segmentation due to poor localization. Prior methods enhance spatial coherence by modifying intermediate attention. But, this coherence isn't consistently propagated to the final output due to subsequent operations such as projections. Additionally, intermediate attention lacks direct interaction with text representations, such semantic discrepancy limits the full potential of CLIP.

In this work, we propose a training-free, feedback-driven self-adaptive framework that adapts output-based patch-level correspondences back to the intermediate attention. The output predictions, being the culmination of the model's processing, encapsulate the most comprehensive visual and textual semantics about each patch. Our approach enhances semantic consistency between internal representations and final predictions by leveraging the model's outputs as a stronger spatial coherence prior. We design key modules, including attention isolation,  confidence-based pruning for sparse adaptation, and adaptation ensemble, to effectively feedback the output coherence cues. Our method functions as a plug-in module, seamlessly integrating into four state-of-the-art approaches with three backbones (ViT-B, ViT-L, ViT-H). We further validate our framework across multiple attention types (Q-K, self-self, and Proxy augmented with MAE, SAM, and DINO). Our approach consistently improves their performance across eight benchmarks. 

% Our method serves as a plugin module, we integrate it into 4 state-of-the-art methods and also validate our framework on various attention types with 3 backbones (ViT-B, ViT-L, ViT-H), incorporating Q-K, self-self, and Proxy attentions with VFMs such as MAE, SAM, and DINO. Our method consistently improves across 8 standard benchmarks.

\end{abstract}    
\section{Introduction}

Open-vocabulary semantic segmentation seeks to localize segments and align visual features with novel class descriptions~\cite{xu2022groupvit,cha2023learning,wu2023metagcd}. CLIP models~\cite{cherti2023reproducible,radford2021learning}, trained on web-scale datasets~\cite{schuhmann2022laion,xu2023demystifying}, have demonstrated remarkable zero-shot performance in open-vocabulary tasks due to strong visual-text alignment~\cite{yu2022coca,li2022blip,radford2021learning}. However, adapting CLIP for semantic segmentation remains challenging~\cite{zhou2022extract} since its pretraining primarily contrasts global image-text representations, limiting fine-grained localization and causing spatial misalignment in patch-level features~\cite{wang2023sclip,lan2024proxyclip}.

\begin{figure}[t!]
    \centering
    \includegraphics[width =\linewidth]{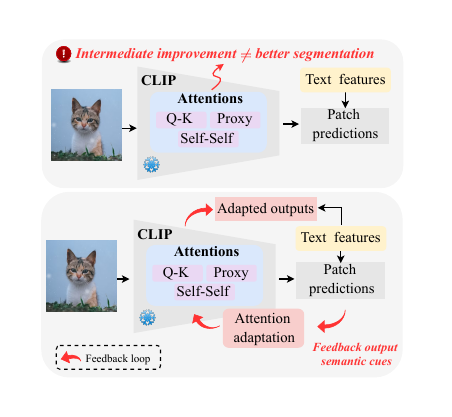} 
    \vspace{-15pt}
    \caption{\textbf{Comparison with the existing training-free methods.} 
    \textbf{Top:} Prior works refine \textit{intermediate} attention but face limitations: (1) improvements may not propagate to final segmentation; (2) attention lacks direct class information. \textbf{Bottom:} We introduce a feedback-driven self-adaptive mechanism that reintegrates semantic outputs with visual and textual cues into CLIP. Our approach is orthogonal to existing methods and serves as a plug-in.
}
\vspace{-15pt}
    \label{fig:illustration}
\end{figure}

Previous efforts to enhance visual localization have primarily focused on fine-tuning CLIP for segmentation tasks~\cite{mukhoti2023open,chen2023exploring,wysoczanska2024clip,wang2023samclip,yuan2024open}. However, fine-tuning compromises CLIP's robust generalization~\cite{wortsman2022robust,zhou2022extract,chi2025learning,chiadapting}. Consequently, recent studies have shifted toward a \textit{training-free} paradigm, addressing the factors contributing to localization degradation in CLIP. ClearCLIP~\cite{lan2024clearclip} identifies residual connections as a major source of noisy segmentation, while SCLIP~\cite{wang2023sclip} highlights the role of self-attention in disrupting spatial information arrangement. Since attention maps capture relationships among patches, patches with similar semantics should be strongly attended to reinforce spatial coherence. Building on this idea, recent works have modulated attention mechanisms, such as replacing Q-K attention with self-self~\cite{wang2023sclip,li2023clip,bousselham2023grounding,lan2024clearclip} or Proxy attentions~\cite{lan2024proxyclip}, both demonstrating improved intermediate spatial consistency.

\begin{figure}[t!]
    \centering
    % \begin{subfigure}{0.45\linewidth}
    % \includegraphics[width =\linewidth]{final_ims/old.png}
    % \caption{Attention map of ProxyCLIP}
    % \end{subfigure}
    % \begin{subfigure}{0.45\linewidth}
    % \includegraphics[width =\linewidth]{final_ims/Front_Proxy_s.png}
    % \caption{Segmentation of ProxyCLIP}
    % \end{subfigure}

    % \begin{subfigure}{0.45\linewidth}
    % \includegraphics[width =\linewidth]{final_ims/ens.png}
    % \caption{Our self-adapted attention}
    % \end{subfigure}
    % \begin{subfigure}{0.45\linewidth}
    % \includegraphics[width =\linewidth]{final_ims/Front_Ours_s.png}
    % \caption{Result of ProxyCLIP + FSA}
    % \end{subfigure}

    % \begin{subfigure}{0.242\linewidth}
    % \captionsetup{labelformat=empty}
    % \includegraphics[width =\linewidth]{ICCV_figs/front/old.png}
    % \caption{Atn. of \cite{lan2024proxyclip}}
    % \end{subfigure}
    % \begin{subfigure}{0.242\linewidth}
    % \captionsetup{labelformat=empty}
    % \includegraphics[width =\linewidth]{ICCV_figs/front/Front_Proxy.png}
    % \caption{Seg. of \cite{lan2024proxyclip}}
    % \end{subfigure}
    % \begin{subfigure}{0.242\linewidth}
    % \captionsetup{labelformat=empty}
    % \includegraphics[width =\linewidth]{ICCV_figs/front/ens.png}
    % \caption{Atn. of \cite{lan2024proxyclip}+FSA}
    % \end{subfigure}
    % \begin{subfigure}{0.242\linewidth}
    % \captionsetup{labelformat=empty}
    % \includegraphics[width =\linewidth]{ICCV_figs/front/Front_ours.png}
    % \caption{Seg. of \cite{lan2024proxyclip}+FSA}
    % \end{subfigure}

 \begin{subfigure}{0.242\linewidth}
    \centering
    \captionsetup{justification=centering}
    \includegraphics[width =\linewidth]{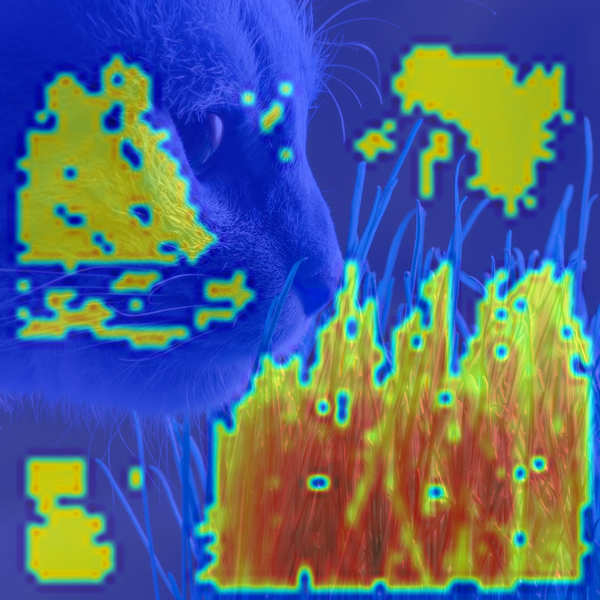}
    \caption{\\ Atn of \cite{lan2024proxyclip}}
\end{subfigure}
\begin{subfigure}{0.242\linewidth}
    \centering
    \captionsetup{justification=centering}
    \includegraphics[width =\linewidth]{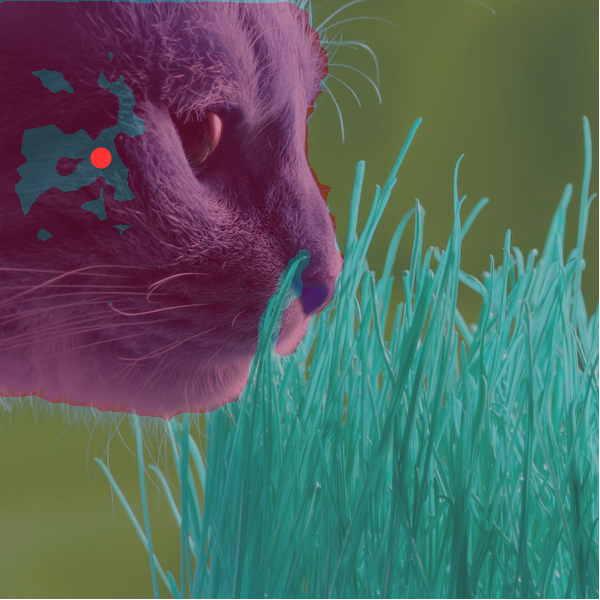}
    \caption{\\ Seg of \cite{lan2024proxyclip}}
\end{subfigure}
\begin{subfigure}{0.241\linewidth}
    \centering
    \captionsetup{justification=centering}
    \includegraphics[width =\linewidth]{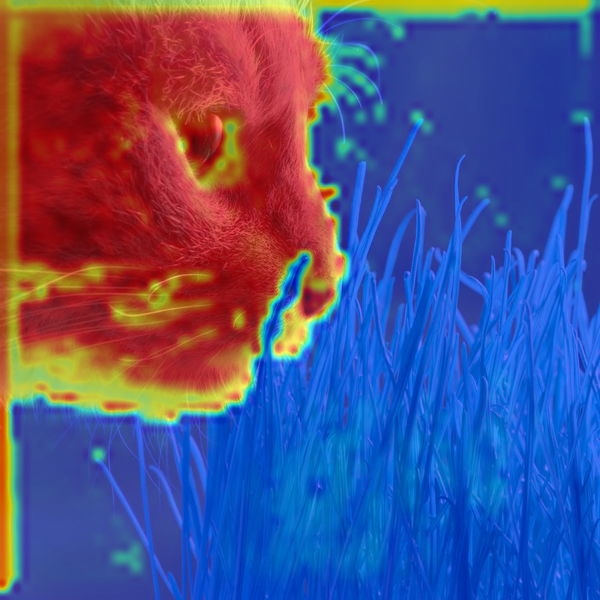}
    \caption{\\Atn of \cite{lan2024proxyclip}+FSA}
\end{subfigure}
\begin{subfigure}{0.242\linewidth}
    \centering
    \captionsetup{justification=centering}
    \includegraphics[width =\linewidth]{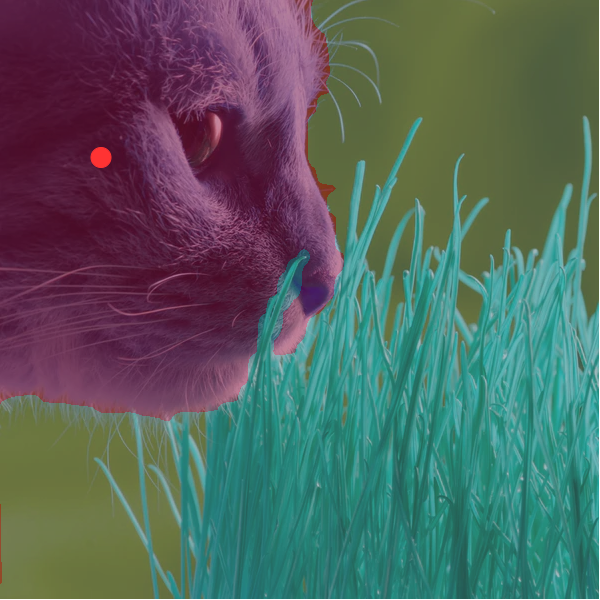}
    \caption{(e) \\ Seg. of \cite{lan2024proxyclip}+FSA}  % Changed (d) to (e)
\end{subfigure}
\vspace{-17pt}
    \caption{
    \textbf{Effectiveness of our feedback self-adaptive mechanism on an image segmented into \textit{grass} and \textit{cat}.} The attention maps correspond to a reference patch (red dot) on the cat face. In (a), the reference patch misaligns with \textit{grass} in the attention map, causing incorrect segmentation in (b). Our proposed FSA adjusts (a) using output predictions to generate semantically aligned attention, focusing on \textit{cat} in (c), and corrects segmentation in (d).
    }
    \vspace{-10pt}
    \label{fig:front}
\end{figure}

However, the final segmentation relies on the output predictions rather than the intermediate attention. Enhanced intermediate coherence may deteriorate due to subsequent operations and hence, might not translate into improved segmentation results. More importantly, intermediate attention does not interface directly with class information. Therefore, neglecting the crucial interplay between intermediate attention and output predictions incorporating text representation is inherently sub-optimal. The output predictions intuitively represent the model's most refined patch-level semantic understanding, encapsulating high-level visual and textual features. We posit that patches with similar predictions likely belong to the same or related class. Leveraging the semantic coherence extracted from the output prediction can provide valuable feedback for further spatial information rearrangement refinement.

Building upon this idea, we propose a \textit{training-free} \textbf{F}eedback-driven \textbf{S}elf-adaptive \textbf{A}ttention (FSA) framework that adapts output semantic cues back into CLIP through a feedback loop. Specifically, we compute pairwise patch similarities between their class predictions, which provides a stronger prior for guiding intermediate attention than using intermediate representations alone. Modulating the original attention maps with these output-based similarities improves consistency between internal representations and final predictions, enhancing information aggregation among semantically similar patches and improving segmentation accuracy. Fig.~\ref{fig:illustration} shows a high-level comparison with the existing methods. The visual example in Fig.~\ref{fig:front} demonstrates that our FSA successfully incorporates the semantic cues at the output to correct the wrongly segmented patches. Our FSA is an unsupervised adaptation method~\cite{liang2022self,caron2021emerging} that leverages the model's self-outputs, sharing similarities with learning paradigms such as test-time adaptation (TTA~\cite{wang2020tent,liu2023meta,sun2020test}) and knowledge distillation (KD~\cite{hinton2015distilling,zhang2021self,zhong2022meta,yebayes,yang2024markov}), as shown in Table~\ref{tab:unsupervised}.

% Our approach is conceptually related to entropy-based test-time adaptation~\cite{wang2020tent,niu2022efficient,song2023ecotta}, which updates parameters during inference by minimizing the entropy of output predictions. However, our feedback self-adaptation is gradient-free, without altering the model parameters. 

To focus on the modulation only on the initial intermediate attention maps, we introduce attention isolation to ensure that the initial output predictions reflect the influence of the original attention rather than interference from subsequent operations. Additionally, we propose confidence-based pruning to filter out irrelevant patches while enhancing the impact of semantically relevant regions. Furthermore, we design three adaptation strategies and integrate them into an ensemble to achieve consistent improvements across diverse attention configurations and backbones.

\begin{table}[t]
    \centering
    % \small
    \scriptsize
    % \footnotesize
    % \vspace{-10pt}
    % \tiny
    \begin{tabular}{c|c|c}
        \toprule
         TTA & KD  & Our feedback mechanism\\
        \hline
        \makecell{Minimizes entropy \\ using output logits.} & \makecell{Treats output logits \\ as soft labels.} & \makecell{Computes patch semantic \\similarity from output logits.} \\
        \bottomrule
    \end{tabular}
    \caption{\textbf{Similarity of our FSA with other learning paradigms.}}
    \label{tab:unsupervised}
    \vspace{-10pt}
\end{table}

% To demonstrate the versatility of our FSA framework, we integrate it with four SoTA methods: MaskCLIP~\cite{zhou2022extract}, SCLIP~\cite{wang2023sclip}, ClearCLIP~\cite{lan2024clearclip} and ProxyCLIP~\cite{lan2024proxyclip}. We further validate on various attention types, including Q-K, self-self, and ProxyCLIP augmented with various VFMs (MAE~\cite{he2022masked}, SAM~\cite{kirillov2023segment}, DINO~\cite{caron2021emerging,oquab2023dinov2}). Evaluated on 8 open-vocabulary segmentation benchmarks, our framework consistently improves across these configurations. Our code for reproducibility will be available upon paper acceptance.  

To demonstrate the versatility of our FSA framework, we integrate it into four SoTA methods: MaskCLIP~\cite{zhou2022extract}, SCLIP~\cite{wang2023sclip}, ClearCLIP~\cite{lan2024clearclip}, and ProxyCLIP~\cite{lan2024proxyclip}. Additionally, we validate FSA across various attention types, including Q-K, self-self, and Proxy attention, augmented  by MAE~\cite{he2022masked}, SAM~\cite{kirillov2023segment}, and DINO~\cite{caron2021emerging,oquab2023dinov2}. Evaluated on eight open-vocabulary segmentation benchmarks, our FSA consistently enhances performance across these configurations. 
Our contributions are summarized as:

% \begin{itemize}
% \item We propose a training-free feedback self-adaptive attention that refines the attention mechanism by adapting output semantic coherence cues back into the model.
% \item We design essential modules, including attention isolation, sparse attention via confidence-based pruning, and adaptation ensemble, to effectively deliver the semantic coherence cues yielded at the output.
% \item Our FSA serves as a plug-in module to enhance various existing methods and various attention configurations. Validation on eight benchmarks demonstrates consistent improvement.
% \end{itemize}

\begin{itemize}
\item We introduce a feedback-driven self-adaptive mechanism that refines the attention process by integrating output-derived semantic coherence cues back into the model.
\item We develop key modules, including attention isolation, sparse attention via confidence-based pruning, and adaptation ensemble, to effectively propagate semantic coherence cues from the output.
\item Our FSA is training-free, and its adaptation process preserves the original model parameters.
\item Our FSA functions as a plug-in module that enhances existing methods and attention configurations, demonstrating effectiveness across eight benchmarks.
\end{itemize}

\section{Related work}

\noindent \textbf{Open-vocabulary segmentation.} CLIP's large-scale pretraining has enabled strong zero-shot transfer for open-vocabulary semantic segmentation \cite{jia2021scaling, khan2022weakly, yu2022coca, li2022blip}, but its global-level pretraining leads to patch-level localization issues \cite{zhou2022extract}. Fine-tuning on segmentation datasets helps address this by enabling local-to-global alignment, often using self-distillation \cite{chen2023exploring, liang2023open, wu2023clipself} or integrating spatial coherence from stronger VFMs into CLIP \cite{wang2023samclip, wysoczanska2024clip, yuan2024open}. However, fine-tuning can reduce CLIP's robustness \cite{wortsman2022robust}, prompting a shift to training-free approaches that modulate problematic components of CLIP for improved performance. For example, ClearCLIP \cite{lan2024clearclip} improves segmentation by altering connections in the last attention block, while MaskCLIP \cite{zhou2022extract} enhances localization with the value embedding. Methods like GEM \cite{bousselham2023grounding}, CLIPSurgery \cite{li2023clip}, and SCLIP \cite{wang2023sclip} replace the original Q-K attention with self-self attentions. ProxyCLIP \cite{lan2024proxyclip} combines CLIP and VFMs, leveraging spatial feature correspondence for state-of-the-art performance. However, most approaches focus on modulating the forward pass, neglecting spatial consistency and class information encoded in the output predictions.

\noindent \textbf{Output-based knowledge transfer.} Output predictions have been effectively used across various machine learning areas, including self-distillation, model compression, domain adaptation, self-supervised learning, uncertainty estimation, and iterative refinement. Self-distillation leverages predictions for intra-model knowledge transfer~\cite{liu2022few}, enhancing generalization and compression~\cite{hinton2015distilling,zhang2019your,hamidi2024train}. In domain and test-time adaptation, output entropy is minimized to to adapt models dynamically without additional labeling~\cite{wang2021tent,wang2024distribution}. Self-supervised learning employs predictions to create pseudo-labels or enforce consistency across augmented views, refining representations without labeled data~\cite{grill2020bootstrap, chen2020simple,chi2021test}. For uncertainty estimation and model calibration, analyzing prediction logits helps assess prediction confidence, with techniques like temperature scaling enhancing reliability~\cite{guo2017calibration,yang2024conditional,yetowards}. Iterative refinement in vision tasks uses predictions to progressively improve predictions, boosting spatial consistency and segmentation accuracy~\cite{lin2017refinenet, cai2018cascade,chi2020all}. These approaches highlight the versatility of output logits for adaptation, optimization, and feedback across diverse tasks. In this work, we introduce a new training-free self-adaptation method that utilizes patch similarity in the output to reorganize the spatial correspondence of intermediate attentions.
\section{Preliminaries}
\noindent \textbf{Segmentation using CLIP ViT.} ViT is composed of attention blocks~\cite{dosovitskiy2020image}. 
% Following the SoTA methods, ProxyCLIP~\cite{lan2024proxyclip} and ClearCLIP~\cite{lan2024clearclip}, we focus exclusively on the \textit{last} attention block, omitting its residual connection and feedforward network. 
Let $\boldsymbol{x} \in \mathbb{R}^{L \times v} = [x_1, \ldots, x_L]^T$ denote the patch tokens, where $L$ is the number of patches and $v$ is the visual feature dimension. Note, we drop \texttt{CLS} token and layer norm~\cite{ba2016layer} for simplicity and follow ProxyCLIP~\cite{lan2024proxyclip} and ClearCLIP~\cite{lan2024clearclip} to modulate the last layer. The resulting visual dense patch features are then obtained by:
\begin{gather}
\boldsymbol{Q} = \boldsymbol{x} \cdot W_Q, \quad \boldsymbol{K} = \boldsymbol{x} \cdot W_K, \quad \boldsymbol{V} = \boldsymbol{x} \cdot W_V,\\
\boldsymbol{z} = \boldsymbol{y'} + \text{FFN}(\boldsymbol{y'}),  \quad \boldsymbol{y'} = \boldsymbol{x} + \text{Proj}(\text{Attn}^{init} \cdot \boldsymbol{V}) \label{eq:attn_block} \\ 
\boldsymbol{Z}^{dense} \in \mathbb{R}^{L\times d} = \mathcal{P}(\boldsymbol{z}),
\end{gather}
where $W_Q, W_K, W_V, \text{Proj}$ are the projection matrices, FFN is the feed-forward network. $\mathcal{P}$ projects the visual features with dimension $v$ to the visual-text shared space with dimension $d$, ($\cdot$) is the matrix multiplication, $\text{Attn}^{init} \in \mathbb{R}^{L \times L}$ represents the \textit{initial} attention maps, which can be the default attention ($\textup{SoftMax}\left(\frac{QK^T}{\tau}\right)$, where $\tau$ is the scaling factor) or other attention configurations, such as self-self ($Q$-$Q$, $K$-$K$, $V$-$V$)~\cite{liu2023grounding,wang2023sclip} and Proxy~\cite{lan2024proxyclip} attention. Note that, in certain methods, e.g., ProxyCLIP and ClearCLIP, the residual and FFN in Eq.~\ref{eq:attn_block} are omitted in the final layer.

To obtain the text representation $\boldsymbol{T} \in \mathbb{R}^{c \times d}$, we input a text prompt, \texttt{"A photo of a [CLASS]"}, combined with $c$ class names into the text encoder. Finally, the patch predictions are computed by taking the cosine similarity:
\begin{equation}
    \boldsymbol{Y}^{dense}\in \mathbb{R}^{L\times c} = \text{cos}(\boldsymbol{Z}^{dense}, \boldsymbol{T}).
\label{eq:logits}
\end{equation}
The segmentation map $\mathcal{M} \in \mathbb{R}^{L \times 1}$ is obtained by $\arg\max$ along the $c$ dimension of $\boldsymbol{Y}^{\text{dense}}$.
\begin{figure}[t!]
    \centering
    \includegraphics[width =\linewidth]{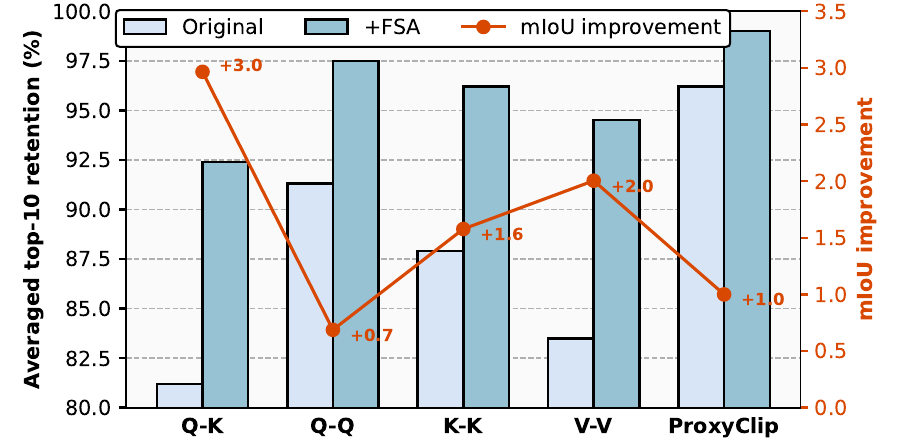} 
    \caption{\textbf{Semantic coherence retention between attention and final predictions (higher is better) and mIoU improvements averaged on 8 benchmarks.} Our self-adaptive attention feeds the patch-wise semantics at output back to modulate the intermediate attention to both improve semantic retention and final mIoU.}
    \label{fig:consistency}
    \vspace{-10pt}
\end{figure}

% \noindent \textbf{Motivation.} Despite that CLIP has greatly advanced in open-vocabulary tasks~\cite{cherti2023reproducible,radford2021learning}, it is less than satisfactory for segmentation tasks due to limited localization capabilities.~\cite{zhou2022extract}. To address this, recent studies have focused on modulating attention to better capture spatial relationships among patches. Empirical results suggest that using self-self or Proxy attention~\cite{lan2024proxyclip} strengthens local feature correspondence, improving segmentation. Yet, segmentation ultimately relies on output logits, not intermediate attention maps. Spatial coherence achieved in attention can be disrupted by subsequent non-linear operations and interaction with text representations, leading to sub-optimal segmentation outcomes.

% \noindent \textbf{Motivation.} CLIP presents insufficient localization capabilities for segmentation tasks~\cite{zhou2022extract}. As the attention is responsible for arranging spatial information~\cite{wang2023sclip}, recent studies have empirically found that self-self~\cite{liu2023grounding,wang2023sclip} or Proxy attention~\cite{lan2024proxyclip} exhibit stronger local feature correspondence. However, segmentation ultimately relies on output predictions rather than intermediate attention. The spatial coherence achieved in $\text{Attn}^{init}$ can be disrupted by subsequent operations, which cannot be fully converyed to the final output.

\noindent \textbf{Motivation.}  CLIP demonstrates limited localization for segmentation~\cite{zhou2022extract}. Since attention governs the arrangement of spatial information~\cite{wang2023sclip}, recent studies have empirically shown that self-self~\cite{liu2023grounding,wang2023sclip} and Proxy attention~\cite{lan2024proxyclip} exhibit stronger local feature correspondence. However, segmentation ultimately depends on output predictions rather than intermediate attention. The spatial coherence established in $\text{Attn}^{init}$ may be disrupted by subsequent operations, preventing it from being fully conveyed to the final output.

\begin{figure*}[t!]
    \centering
    \includegraphics[width =\linewidth]{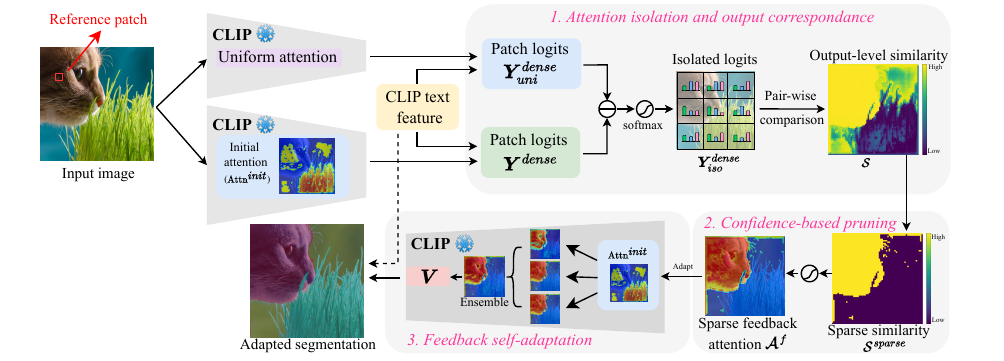} 
    % {final_ims/overview.pdf} 
    \caption{\textbf{Overview of our feedback-driven self-adaptive attention framework.} 
    We isolate the contribution of initial attention using a dual branch with uniform attention. Pairwise patch similarity is then calculated by applying KL divergence to the output logits. A confidence-based pruning step follows to eliminate irrelevant patches and amplify the correlation between related patches. The resulting sparse feedback attention is adapted back to the initial attention maps using three operations, which are subsequently ensembled.}
    \label{fig:overview}
    \vspace{-10pt}
\end{figure*}

We first examine such spatial coherence retention by comparing the intermediate attention $\text{Attn}^{init}$ (final layer as in ProxyCLIP) and the final segmentation map $\mathcal{M}$. Intuitively, \textit{if two patches have high mutual attention scores, they should be ideally predicted to the same class}. For the $i^{th}$ patch, we identify the $k$ patches with Top-\textit{k} score in the $i^{th}$ row of $\text{Attn}^{init}$ and check whether \textit{any} of the $k$ patches is predicted as the same class of $i^{th}$ patch in $\mathcal{M}$:
\begin{equation}
\text{Retention} = \frac{1}{L} \sum_{i=1}^{L} \max_{j \in \text{Top-}k(\text{Attn}^{init}_{i})} \mathbb{I}(\mathcal{M}_i = \mathcal{M}_j),
\label{eq:consistency}
\end{equation}
where $\mathbb{I}$ is an indicator that equals 1 if the condition is met. To avoid self-comparisons, we set the diagonal elements ($\text{Attn}^{init}_{ii})$ to zero. Fig.~\ref{fig:consistency} shows the average Top-$10$ retention across 8 benchmarks for various attention configurations. Obviously, such spatial coherence is not fully retained in the final predictions. Furthermore, the attention map does not interface with class information directly. In contrast, output predictions capture comprehensive visual semantics and integrate class information from textual features. This motivates our feedback self-adaptive approach, which leverages these outputs as semantic cues to rearrange and modulate spatial information in the intermediate attention $\text{Attn}^{init}$. Our method enhances semantic consistency and segmentation performance, as demonstrated in Fig.~\ref{fig:consistency}. In Appendix C, we demonstrate another metric and show the retention degradation after each operation.

\section{Method}

% \noindent In this section, we introduce the feedback self-adaptive attention (FSA) framework. We first build effective self-adaptive attention by global bias removal (Sec.\ref{sec:debias}) and confidence-based pruning (Sec.\ref{sec:pruning}). We then feed the self-adaptive attention back to modulate the initial attention the $\text{Attn}^{init}$ (Sec.\ref{sec:adaptation}).

% \noindent  In this section, we present the feedback self-adaptive attention (FSA) framework as in Fig.~\ref{fig:overview}. We first construct effective self-adaptive attention through global bias removal (Sec.\ref{sec:debias}) and confidence-based pruning (Sec.\ref{sec:pruning}). Then, we adapt the self-adaptive attention back to modulate the initial attention $\text{Attn}^{init}$ to rearrange the spatial information (Sec.\ref{sec:adaptation}).

\noindent  In this section, we present the feedback self-adaptive attention (FSA) framework (Fig.~\ref{fig:overview}). We first construct effective self-adaptive attention through attention isolation (Sec.\ref{sec:debias}) and confidence-based pruning (Sec.\ref{sec:pruning}). Then, we adapt the self-adaptive attention to modulate the spatial information in the initial attention $\text{Attn}^{init}$ (Sec.\ref{sec:adaptation}).

% \yang{this section reads very dry (only text and math). Can we add some figures to visualize (e.g. visualizing the attention or similarity) what happens after each subsection? Similar to Fig 1, but for each step}

\subsection{Stand-alone intermediate attention isolation}\label{sec:debias}

% CLIP may exhibit intrinsic global biases, such as preferences for specific classes or features~\cite{wolfe2022markedness,tanjim2024discovering}, which can affect the output logits regardless of the attention patterns used. As we aim to only adapt $\text{Attn}^{init}$ for spatial knowledge rearrangement, it is essential that the output logits reflect only the semantic information induced by the initial attention. Please note that such bias is temporarily suppressed and will be added back inherently as we only modulate the original attention, keeping other operations untouched.

% The attention map $\text{Attn}^{init}$ is the source of spatial coherence measurement to control which and how intensive the correlated patches should be attended. Therefore, we aim to only modulate $\text{Attn}^{init}$ for spatial knowledge rearrangement. However, As shown in Eq.~\ref{eq:attn_block}-\ref{eq:logits}, the logits are obtained following the process of $\text{Attn}^{init} \cdot \boldsymbol{V} \scriptstyle{\rightarrow}$ sub-modules (projections, FFN, etc.) $\scriptstyle{\rightarrow}$ text $\scriptstyle{\rightarrow}$ logits. Obviously, logits $\boldsymbol{Y}^{dense}$ do not purely reflect the stand-alone contribution from $\text{Attn}^{init}$ which serves as an intermediate measurement. 

The initial attention map, $\text{Attn}^{init}$, serves as the foundation for spatial coherence, determining both the selection of correlated patches and the intensity of their attention. Thus, our objective is to modulate $\text{Attn}^{init}$ solely for spatial knowledge rearrangement. However, as shown in Eq.~\ref{eq:attn_block}-\ref{eq:logits}, the logits are derived through a sequential process: $\text{Attn}^{init} \cdot \boldsymbol{V} \scriptstyle{\rightarrow}$ sub-modules (projections, FFN, etc.) $\scriptstyle{\rightarrow}$ text $\scriptstyle{\rightarrow}$ logits. Consequently, the dense logits, $\boldsymbol{Y}^{dense}$, do not exclusively reflect the independent contribution of $\text{Attn}^{init}$, as it functions merely as an intermediate measure within this pipeline.

To address the interference, we devise an isolation procedure that preserves the stand-alone contribution of \(\text{Attn}^{init}\) in \(\boldsymbol{Y}^{dense}\), making it crucial that output relationships reflect only the initial attention maps. Specifically, we attach a parallel branch with a manually crafted \textit{uniform} attention map \(\text{Attn}^{uni} = \tfrac{1}{L}\mathbf{1}_{L\times L}\), ensuring every patch is equally attended. The resulting logits \(\boldsymbol{Y}^{dense}_{uni}\) are obtained following the same steps in Eqs.~\ref{eq:attn_block}--\ref{eq:logits}. Both \(\text{Attn}^{uni}\) and \(\text{Attn}^{init}\) traverse identical sub-modules and text-alignment stages, allowing us to isolate the learned attention’s net effect by subtracting the uniform-based logits from the original. Therefore, we can pinpoint how much of the final logits is genuinely driven by the learned selective attention patterns rather than a globally ``spread out'' context. The resulting subtraction is then transformed into a probability distribution via softmax:
\begin{equation}
    \boldsymbol{Y}^{dense}_{iso} = \text{Softmax}\bigl(\boldsymbol{Y}^{dense} - \boldsymbol{Y}^{dense}_{uni}\bigr).
\end{equation}

We then quantify the relationships between patches by calculating pair-wise KL divergence:
\begin{equation}
    \mathcal{D}(i, j) = \text{KL}(\boldsymbol{Y}^{dense}_{iso, i} \parallel \boldsymbol{Y}^{dense}_{iso, j}),
\end{equation}
where $\mathcal{D} \in \mathbb{R}^{L \times L}$ records the divergence for each pair of patches. To emphasize patch correspondence rather than difference, we convert the divergence to similarity as:
\begin{equation}
    \mathcal{S} \in \mathbb{R}^{L \times L}=  \frac{1}{\mathcal{D} + 1}.
\label{eq:similarity}
\end{equation}
The KL divergence, bounded in $[0, -\infty]$, is mapped to $[0, 1]$, with a score of 1 indicating maximum similarity. When attention isolation is applied, the output logits align better with the intermediate attention, leading to further segmentation gains.

\subsection{Confidence-based sparse attention}\label{sec:pruning}

The similarity map $\mathcal{S}$ not only encodes visual semantic coherence among patches but more importantly, it integrates \textit{class cues} from text representations. Capturing each patch’s response to each class offers a robust foundation for adapting attention with contextually relevant cues. However, in segmentation tasks, patches may belong to various classes, each with unique semantic information. Since $\mathcal{S}$ contains similarities for every patch pair, low-correlated patches can dilute the relationships between strongly correlated patches. Thus, it is effective to suppress irrelevant patches while amplifying semantically coherent ones.

To this end, we propose a confidence-based sparse attention. Specifically, we first normalize each row of $\mathcal{S}$ using softmax to obtain $\mathcal{\hat{S}}$, resulting in a confidence distribution across patches. Let $\mathcal{\hat{S}}_i$ denote the $i^{\text{th}}$ row which represents the confidence of similarity between the $i^{\text{th}}$ patch and all other patches. We sort $\mathcal{\hat{S}}_i$ in descending order ($\mathcal{\hat{S}}^{\text{sorted}}_i$) and record the sorting indices ($idx$). For each position, we compute the \textit{cumulative} confidence probability:
\begin{equation}
    \mathcal{C}_{i,j}^{\text{sorted}} = \sum_{k=1}^{j} \mathcal{\hat{S}}^{\text{sorted}}_{i,k}.
    \label{eq:cumulative}
\end{equation}
We then revert to their original order using $idx$: $\mathcal{C}_i = \mathcal{C}_i^{\text{sorted}}[idx]$. Repeating for each row yields the final cumulative confidence map $\mathcal{C}$. We then set a confidence level $p$ to selectively amplify higher similarity values with adaptive exponential scaling and fully suppress the others:
\begin{equation}
    \mathcal{S}^{sparse}_{i,j} = 
\begin{cases} 
      \mathcal{S}_{i,j} \cdot  e^{\lambda \cdot \mathcal{S}_{i,j}}, & \text{if } \mathcal{C}_{i,j} \leq p \text{ or } j = 1 \\
      -\infty, & \text{otherwise},
      \label{eq:scale}
   \end{cases}
\end{equation}
where $\lambda$ and $p$ are the only hyperparameters in our method, controlling the scaling sharpness and the cutoff confidence, set empirically to 2 and 0.45, respectively. Finally, we convert the sparse similarity map into our feedback attention: $\mathcal{A}^f = \text{softmax}(\mathcal{S}^{sparse})$. 

In summary, our confidence-based sparse attention (Eq.~\ref{eq:cumulative} \& \ref{eq:scale}) can be interpreted as: \textit{selecting patches starting with the highest confidence until cumulative confidence of selected patches reaches $p$}. It emphasizes semantically coherent patches while ignoring low-confidence ones.
A pseudo-code can be found in the supplementary.

\subsection{Feedback self-adaptive attention}\label{sec:adaptation}
The resulting attention $\mathcal{A}^f$ captures the ultimate semantic relationships between pairs of patches, derived from the initial attention. We reintegrate this feedback into the original pipeline (Eq.~\ref{eq:attn_block}-\ref{eq:logits}) to enhance the spatial arrangement of information. Specifically, we propose three \textit{training-free} adaptations to modify the interaction with $\boldsymbol{V}$ in Eq.~\ref{eq:attn_block}:
% \begin{numcases}{\text{Attn}^{init} \cdot \boldsymbol{V} \xrightarrow{\text{adaptation}}}
%    \mathcal{A}^f \cdot (\text{Attn}^{init} \cdot \boldsymbol{V}), \label{eq:1} \\
%    \text{Attn}^{init} \cdot (\mathcal{A}^f \cdot \boldsymbol{V}), \label{eq:2} \\
%    \mathcal{A}^f \cdot \boldsymbol{V}. \label{eq:3}
% \end{numcases}
% Add space before the equation

\vspace{-1em} % Adjust the space size as needed
\begin{numcases}{\text{Attn}^{init} \cdot \boldsymbol{V} \xrightarrow{\text{adaptation}}}
   \mathcal{A}^f \cdot (\text{Attn}^{init} \cdot \boldsymbol{V}), \label{eq:1} \\
   \text{Attn}^{init} \cdot (\mathcal{A}^f \cdot \boldsymbol{V}), \label{eq:2} \\
   \mathcal{A}^f \cdot \boldsymbol{V}. \label{eq:3}
\end{numcases}
\noindent Eq.~\ref{eq:1} captures the $\text{Attn}^{init}$-driven relationships first, then selectively amplifies or suppresses them based on feedback, effectively introducing a two-step refinement. Eq.~\ref{eq:2} immediately influences 
$\boldsymbol{V}$ before initial attention is applied, giving higher priority to semantic coherence and confidence information from the output logits. Eq.~\ref{eq:3} emphasizes only the feedback-driven information solely relying on the output semantic coherence and confidence. We observe that each of the three adaptations offers distinct benefits depending on the backbone, attention configuration, and benchmark. To leverage their combined strengths, we propose an ensemble approach by isolating $\boldsymbol{V}$ in Eq.~\ref{eq:1}-\ref{eq:3}:
\begin{equation}
   (\frac{\mathcal{A}^f \cdot \text{Attn}^{init} + \text{Attn}^{init} \cdot \mathcal{A}^f + \mathcal{A}^f}{3})\cdot \boldsymbol{V}. \label{eq:ens}
\end{equation}
We denote the adapted logits from Eq.~\ref{eq:1}-\ref{eq:ens} as $\boldsymbol{Y}^{dense}_{1}$, $\boldsymbol{Y}^{dense}_{2}$, $\boldsymbol{Y}^{dense}_{3}$, and $\boldsymbol{Y}^{dense}_{ens}$ which are processed using the $\arg\max$ to generate the adapted segmentation map.

\section{Experiments}
In this section, we first validate our FSA framework by integrating it with various SoTA methods and attention configurations. We then conduct comprehensive ablation studies to assess the effectiveness of each proposed component.

% \subsection{Dataset and implementation details}
% \noindent \textbf{Datasets and implementations.} We follow ProxyCLIP~\cite{lan2024proxyclip}, CLIP-DINOiser~\cite{wysoczanska2023clip}, and TCL~\cite{cha2023learning} to evaluate our FSA framework on 8 standard benchmarks: (\textit{i}) with background class: PASCAL VOC \cite{everingham2012pascal} (VOC), PASCAL Context \cite{mottaghi2014role} (Context), and COCO Object \cite{caesar2018coco} (Object); and (\textit{ii}) without background class: PASCAL VOC20 \cite{everingham2012pascal} (VOC20), PASCAL Context59 \cite{mottaghi2014role} (Context59), COCO Stuff \cite{caesar2018coco} (Stuff), Cityscapes \cite{cordts2016cityscapes} (City), and ADE20K \cite{zhou2019semantic} (ADE). We use the ProxyCLIP image preprocessing pipeline, resizing images to a shorter side of 336 pixels for PASCAL and COCO datasets, and 448 pixels for Cityscapes and ADE20K. A sliding window strategy with a $336 \times 336$ window and $112 \times 112$ stride is applied. We adopt the standard ImageNet text prompt template~\cite{radford2021learning} with benchmark-specific class names for text features. All experiments report mean Intersection over Union (mIoU) on the validation sets without training and post-processing. We adopt ProxyCLIP~\cite{lan2024proxyclip} and  MMSegmentation \cite{contributors2020mmsegmentation} codebase.

\subsection{Implementation details and datasets}
\noindent \textbf{Architectures.} To show the versatility of FSA, we test on 3 backbones: ViT-B/16 and ViT-L/14 from CLIP~\cite{radford2021learning}, and ViT-H/14 from OpenCLIP~\cite{cherti2023reproducible}. We adopt DINO ViT-B/8~\cite{caron2021emerging} for ProxyCLIP as default unless specified otherwise.

% \noindent \textbf{Implementations.} Our FSA serves as a plugin module which can be integrated into SoTA methods. We inherent their data processing and only modify the model part. We adopt the standard ImageNet text prompt template~\cite{radford2021learning} with benchmark-specific class names for text features. All experiments report mean Intersection over Union (mIoU) on the validation sets without training and post-processing. We adopt the MMSegmentation \cite{contributors2020mmsegmentation} codebase.

\noindent \textbf{Implementations.} Our FSA functions as a plug-in module that can be integrated into SoTA methods, modifying only the model component. Note that, we preserve their original data processing pipeline. For example patch size and stride size are kept the same. All experiments report mean Intersection over Union (mIoU) on validation sets, without training or post-processing. Implementation is based on the MMSegmentation codebase~\cite{contributors2020mmsegmentation}.

\noindent \textbf{Datasets.} We follow TCL~\cite{cha2023learning}, ProxyCLIP~\cite{lan2024proxyclip},  SCLIP~\cite{wang2023sclip} and CLIP-DINOiser~\cite{wysoczanska2023clip} to evaluate on eight standard benchmarks: (\textit{i}) with background class: PASCAL VOC \cite{everingham2012pascal} (VOC), PASCAL Context \cite{mottaghi2014role} (Context), and COCO Object \cite{caesar2018coco} (Object); and (\textit{ii}) without background class: PASCAL VOC20 \cite{everingham2012pascal} (VOC20), PASCAL Context59 \cite{mottaghi2014role} (Context59), COCO Stuff \cite{caesar2018coco} (Stuff), Cityscapes \cite{cordts2016cityscapes} (City), and ADE20K \cite{zhou2019semantic} (ADE).

\begin{table*}[!t]
  \tabcolsep10pt
  \centering
  % \small
  % \footnotesize
  \scriptsize
  % \multirow{-3}{*}{\shortstack{ViT-L/14}
  % \small
  \begin{tabular}{c|l|ccccccccl}
    \toprule
    Models & Methods  & VOC & Context & Object & VOC20 & Context59 & Stuff & City & ADE & Avg. \\
    \midrule
    % \hline
     & MaskCLIP\textsubscript{\textcolor{gray}{ECCV'22}} \cite{zhou2022extract}& 38.8 & 23.6 & 20.6 & 74.9 & 26.4 & 16.4 & 12.6 & 9.8 & 27.9 \\
    \rowcolor{gray!18.0} \cellcolor{white} & + FSA & 
    47.7&31.0&29.2&78.3&34.2&21.4&28.4&16.0&\bestimprovered{35.8}{7.9} \\
    % \cmidrule(){2-11}
    \hhline{~----------}
    % & \multicolumn{10}{|c}{\hline} \\
    & SCLIP\textsubscript{\textcolor{gray}{ECCV'24}} \cite{wang2023sclip} & 59.1 & 30.4 & 30.5 & 80.4 & 34.2 & 22.4 & 32.2 & 16.1 & 38.2 \\
    \rowcolor{gray!18.0} \cellcolor{white} & + FSA & 
    61.5&33.3&33.9&82.8&36.8&24.4&34.7&17.5&\bestimprovered{40.6}{2.4}\\
    \hhline{~----------}
    & ClearCLIP\textsubscript{\textcolor{gray}{ECCV'24}}~\cite{lan2024clearclip}&51.8	&32.6&	33.0&	80.9&	35.9&	23.9&	30.0&	16.7&	38.1\\
    \rowcolor{gray!18.0} \cellcolor{white} & + FSA & 53.0&	36.6&	33.2&	81.3&	33.8&	24.3&	30.8&	17.4&	\bestimprovered{38.8}{0.7}\\
    \hhline{~----------}
    & ProxyCLIP\textsubscript{\textcolor{gray}{ECCV'24}}~\cite{lan2024proxyclip} & 61.3 & 35.3 & 37.5 & 80.3 & 39.1 & 26.5 & 38.1 & 20.2 & 42.3 \\
     \rowcolor{green!18.0} \cellcolor{white} \multirow{-8}{*}{\shortstack{\makecell{\cellcolor{white}{CLIP}\\ \cellcolor{white}{ViT-B/16}}}} & + FSA (Ours) & 63.7 & 36.1	& 38.0& 82.3	&39.9	&27.0	&38.8	&20.5	&\bestimprovered{43.3}{1.0} \\

    %%% L14
    \midrule
    & MaskCLIP\textsubscript{\textcolor{gray}{ECCV'22}} \cite{zhou2022extract} & 23.3 & 11.7 & 7.2 & 29.4 & 12.4 & 8.8 & 11.5 & 7.2 & 13.9 \\
    \rowcolor{gray!18.0} \cellcolor{white} & + FSA & 44.8&26.8&27.8&73.9&29.4&19.0&23.1&16.2&\bestimprovered{32.6}{18.7} \\

    \hhline{~----------}
    & SCLIP\textsubscript{\textcolor{gray}{ECCV'24}} \cite{wang2023sclip} & 43.5 & 22.3 & 25.0 & 69.1 & 25.2 & 17.6 & 18.6 & 10.9 & 29.0\\
    \rowcolor{gray!18.0} \cellcolor{white} & + FSA &48.1&	27.8&	30.8&	79.9&	30.3&	20.4&	27.1&	15.9&\bestimprovered{35.0}{6.0}\\
    
    \hhline{~----------}
    & ClearCLIP\textsubscript{\textcolor{gray}{ECCV'24}}~\cite{lan2024clearclip}&46.1&	29.6&	26.7&	80.0&	30.1&	19.9&	27.9&	15.0&	34.4\\
    \rowcolor{gray!18.0} \cellcolor{white} & + FSA &47.5&	30.8&	27.9&	80.4&	30.2&	20.4&	27.2&	16.8&\bestimprovered{35.2}{0.8}\\
    
    \hhline{~----------}
    & ProxyCLIP\textsubscript{\textcolor{gray}{ECCV'24}}~\cite{lan2024proxyclip} & 60.6 & 34.5 & 39.2 & 83.2 & 37.7 & 25.6 & 40.1 & 22.6 & 42.9 \\
    \rowcolor{green!18.0} \cellcolor{white} \multirow{-8}{*}{\shortstack{\makecell{\cellcolor{white}{CLIP}\\ \cellcolor{white}{ViT-L/14}}}} & + FSA (Ours) & 61.8	&34.9	&40.2	&84.1	&38.1	&25.9	&41.2 &22.9	&\bestimprovered{43.6}{0.7} \\

    %%% H14
    \midrule
    % \multirow{5}{*}{\makecell{CLIP\\ ViT-H/14}} & OpenCLIP 
    
    & MaskCLIP\textsubscript{\textcolor{gray}{ECCV'22}} \cite{zhou2022extract}  & 31.4 & 13.3 & 16.2 & 41.7 & 15.8 & 8.4 & 17.7 & 10.4 & 19.3  \\
    \rowcolor{gray!18.0} \cellcolor{white} & + FSA & 44.8&	27.7&	27.6&	71.3&	30.5&	20.7&	26.1&	19.0&\bestimprovered{33.4}{14.1}\\

    \hhline{~----------}
    & SCLIP\textsubscript{\textcolor{gray}{ECCV'24}} \cite{wang2023sclip} & 43.8 & 23.5 & 24.6 & 67.5 & 25.6 & 16.8 & 19.5 & 11.3 & 29.1 \\
    \rowcolor{gray!18.0} \cellcolor{white} & + FSA & 48.1&	28.4&	29.1&	76.4&	31.3&	20.6&	29.1&	18.1&\bestimprovered{35.1}{6.0}\\

    \hhline{~----------}
    & ClearCLIP\textsubscript{\textcolor{gray}{ECCV'24}}~\cite{lan2024clearclip} & 46.0&	27.1&	27.9&	78.1&	29.8&	19.3&	28.3&	16.0&	34.1\\
    \rowcolor{gray!18.0} \cellcolor{white} & + FSA & 47.2&	31.1&	28.3&	78.6&	28.5&	20.6&	29.3&	18.7&\bestimprovered{35.3}{1.2}	\\

    \hhline{~----------}
    & ProxyCLIP\textsubscript{\textcolor{gray}{ECCV'24}}~\cite{lan2024proxyclip} & 65.0 & 35.4 & 38.6 & 83.3 & 39.6 & 26.8 & 42.0 & 24.2 & 44.4 \\
    \rowcolor{green!18.0} \cellcolor{white} \multirow{-8}{*}{\shortstack{\makecell{\cellcolor{white}{CLIP}\\ \cellcolor{white}{ViT-H/14}}}}  & +FSA (Ours) & 67.9	& 36.3 &	40.2	& 85.7 &	40.5	&27.3	&43.6	&24.5	&\bestimprovered{45.8}{1.4} \\
    
  \bottomrule
  \end{tabular}
  \caption{\textbf{Integration of FSA into the state-of-the-art methods for open-vocabulary semantic segmentation on eight standard benchmarks.} Our feedback self-adaptive framework consistently improves previous approaches across all datasets and three different backbones.}
  \vspace{-10pt}
  \label{tab:main_results}
\end{table*}

\begin{table}[!t]
  \tabcolsep3.8pt
  \centering
    % \scriptsize
    \tiny
  \begin{tabular}{c|c|l|lllllllll}
    \toprule
    CLIP & VFMs & Methods & V21 & C60 & Obj & V20 & C59 & Stf & City & ADE & Avg.  \\
    \midrule
    % VIT b16 ****************************************
    % \multirow{8}{*}{\makecell{CLIP\\ ViT-B/16}} &

    \multirow{6}{*}{\rotatebox{90}{ViT-B/16}} &
    \multirow{2}{*}{\makecell{SAM \\ ViT-B/16}} & Proxy & 59.3 & 33.6 & 35.4 & 80.4 & 37.0 & 25.0 & 37.0 & 19.1 & 40.8 \\
    && \cellcolor{gray!20.0}+ FSA & \cellcolor{gray!20.0}60.7 & \cellcolor{gray!20.0}34.0 & \cellcolor{gray!20.0}35.8 & \cellcolor{gray!20.0}81.8 & \cellcolor{gray!20.0}37.4 & \cellcolor{gray!20.0}25.2 & \cellcolor{gray!20.0}37.9 & \cellcolor{gray!20.0}19.3 & \cellcolor{gray!20.0}\bestimprovered{41.5}{0.7} \\
    \cmidrule(){2-12}
    & \multirow{2}{*}{\makecell{MAE \\ViT-B/16}} & Proxy & 52.2 & 30.4 & 30.8 & 76.3 & 33.5 & 23.1 & 30.1 & 17.1 & 36.7\\
    & & \cellcolor{gray!20.0}+ FSA & \cellcolor{gray!20.0}54.3 & \cellcolor{gray!20.0}30.9 & \cellcolor{gray!20.0}31.3 & \cellcolor{gray!20.0}78.1 & \cellcolor{gray!20.0}33.9 & \cellcolor{gray!20.0}23.4 & \cellcolor{gray!20.0}33.6 & \cellcolor{gray!20.0}17.5 & \cellcolor{gray!20.0}\bestimprovered{37.9}{1.2}\\
    \cmidrule(){2-12}
    & \multirow{2}{*}{\makecell{DINOv2\\ ViT-B/14}} & Proxy & 58.6 & 33.8 & 37.0 & 83.0 & 37.2 & 25.4 & 33.9 & 19.7 & 41.1 \\
    & & \cellcolor{gray!20.0}+ FSA & \cellcolor{gray!20.0}59.2 & \cellcolor{gray!20.0}33.9 & \cellcolor{gray!20.0}37.4 & \cellcolor{gray!20.0}84.0 & \cellcolor{gray!20.0}37.5 & \cellcolor{gray!20.0}25.5 & \cellcolor{gray!20.0}34.4 & \cellcolor{gray!20.0}19.7 & \cellcolor{gray!20.0}\bestimprovered{41.4}{0.3} \\
    \midrule
    % VIT - L14 **************************************
     \multirow{6}{*}{\rotatebox{90}{ViT-L/14}} &  \multirow{2}{*}{\makecell{SAM \\ ViT-B/16}} & Proxy & 57.2 & 32.6 & 36.5 & 82.3 & 35.6 & 24.2 & 39.1 & 20.7 & 41.0 \\
    && \cellcolor{gray!20.0}+ FSA & \cellcolor{gray!20.0}58.5 & \cellcolor{gray!20.0}33.0 & \cellcolor{gray!20.0}38.0 & \cellcolor{gray!20.0}83.1 & \cellcolor{gray!20.0}36.1 & \cellcolor{gray!20.0}24.5 & \cellcolor{gray!20.0}40.5 & \cellcolor{gray!20.0}21.0 & \cellcolor{gray!20.0}\bestimprovered{41.8}{0.8}\\
    \cmidrule(){2-12}
    & \multirow{2}{*}{\makecell{MAE \\ViT-B/16}} & Proxy & 49.0 & 27.8 & 31.6 & 78.3 & 30.2 & 20.8 & 31.8 & 17.2 & 35.8\\
    & & \cellcolor{gray!20.0}+ FSA & \cellcolor{gray!20.0}52.8 & \cellcolor{gray!20.0}29.9 & \cellcolor{gray!20.0}34.9 & \cellcolor{gray!20.0}80.2 & \cellcolor{gray!20.0}32.5 & \cellcolor{gray!20.0}22.6 & \cellcolor{gray!20.0}34.7 & \cellcolor{gray!20.0}19.0 & \cellcolor{gray!20.0}\bestimprovered{38.3}{2.5}\\
    \cmidrule(){2-12}
    & \multirow{2}{*}{\makecell{DINOv2\\ ViT-B/14}} & Proxy & 56.6 & 33.0 & 36.7 & 85.2 & 36.2 & 24.6 & 35.2 & 21.6 & 41.1 \\
    & & \cellcolor{gray!20.0}+ FSA & \cellcolor{gray!20.0}57.4 & \cellcolor{gray!20.0}33.3 & \cellcolor{gray!20.0}38.0 & \cellcolor{gray!20.0}85.8 & \cellcolor{gray!20.0}36.5 & \cellcolor{gray!20.0}24.7 & \cellcolor{gray!20.0}36.1 & \cellcolor{gray!20.0}21.9 & \cellcolor{gray!20.0}\bestimprovered{41.7}{0.6} \\
    \midrule
    % VIT - H14 **************************************
     \multirow{6}{*}{\rotatebox{90}{ViT-H/14}} &  \multirow{2}{*}{\makecell{SAM \\ ViT-B/16}} & Proxy & 63.5 & 34.1 & 36.7 & 84.0 & 37.9 & 25.0 & 41.1 & 22.0 & 43.1 \\
    && \cellcolor{gray!20.0}+ FSA & \cellcolor{gray!20.0}64.9 & \cellcolor{gray!20.0}34.4 & \cellcolor{gray!20.0}37.6 & \cellcolor{gray!20.0}85.5 & \cellcolor{gray!20.0}38.0 & \cellcolor{gray!20.0}25.1 & \cellcolor{gray!20.0}42.6 & \cellcolor{gray!20.0}21.9 & \cellcolor{gray!20.0}\bestimprovered{43.7}{0.6}\\
    \cmidrule(){2-12}
    & \multirow{2}{*}{\makecell{MAE \\ViT-B/16}} & Proxy & 54.7 & 29.8 & 32.2 & 80.6 & 32.9 & 21.8 & 34.9 & 19.4 & 38.3\\
    & & \cellcolor{gray!20.0}+ FSA  & \cellcolor{gray!20.0}58.7 & \cellcolor{gray!20.0}32.0 & \cellcolor{gray!20.0}35.2 & \cellcolor{gray!20.0}82.2 & \cellcolor{gray!20.0}35.3 & \cellcolor{gray!20.0}24.0 & \cellcolor{gray!20.0}37.9 & \cellcolor{gray!20.0}21.3 & \cellcolor{gray!20.0}\bestimprovered{40.8}{2.5}\\
    \cmidrule(){2-12}
    & \multirow{2}{*}{\makecell{DINOv2\\ ViT-B/14}} & Proxy & 61.5 & 34.0 & 37.3 & 86.1 & 37.8 & 26.2 & 37.8 & 23.4 & 43.0 \\
    & & \cellcolor{gray!20.0}+ FSA & \cellcolor{gray!20.0}63.0 & \cellcolor{gray!20.0}34.4 & \cellcolor{gray!20.0}38.5 & \cellcolor{gray!20.0}87.4 & \cellcolor{gray!20.0}38.4 & \cellcolor{gray!20.0}26.4 & \cellcolor{gray!20.0}39.7 & \cellcolor{gray!20.0}23.7 & \cellcolor{gray!20.0}\bestimprovered{43.9}{0.9} \\
  \bottomrule
  \end{tabular}
  \caption{\textbf{Integration of our FSA into ProxyCLIP with various VFMs.} Consistent improvements across VFMs, benchmarks, and backbones highlight the versatility and effectiveness of our FSA.
  }
  \vspace{-10pt}
\label{tab:VFM_attn}
\end{table}

\subsection{Main results}
\noindent\textbf{Integration into the SoTA methods.} Table~\ref{tab:main_results} reports the integration of our FSA with training-free SoTA methods including: 
% SegCLIP \cite{luo2023segclip},
% ViewCo \cite{ren2023viewco},
% OVSegmentor \cite{xu2023learning},
% CoCu \cite{xing2023rewrite}, 
% SAM-CLIP \cite{wang2023samclip},
% GroupViT \cite{xu2022groupvit}, 
% TCL \cite{cha2023learning},  and CLIP-DINOiser \cite{wysoczanska2023clip}. and training-free:
% CLIPSurgery \cite{li2023clip},
% GEM \cite{bousselham2023grounding},
% ReCo \cite{shin2022reco},
MaskCLIP \cite{zhou2022extract}, 
SCLIP \cite{wang2023sclip},
ClearCLIP \cite{lan2024clearclip}
and ProxyCLIP~\cite{lan2024proxyclip}. 
Our method consistently improves all of them across all eight benchmarks. Notably, we surpass MaskCLIP by a significant margin of \textbf{+7.9 mIoU}, \textbf{+18.7 mIoU} and \textbf{+14.1 mIoU} using ViT-B/16, L/14, and H/14, respectively. Obvious improvement is also observed on SCLIP. ProxyCLIP leverages external knowledge from VFMs, providing strong spatial coherence. However, our FSA system further enhances performance across diverse CLIP architectures, pushing their boundaries.
% Built on ProxyCLIP, our approach leverages a feedback self-adaptive system that integrates semantic coherence from the output to correct misaligned spatial patches. Consequently, we achieve consistent improvements over ProxyCLIP, with average gains of \textbf{+1.0 mIoU}, \textbf{+0.7 mIoU}, and \textbf{+1.4 mIoU} using ViT-B/16, ViT-L/14, and ViT-H/14 backbones, respectively. Notably, our results on ViT-B/16 surpass those of ProxyCLIP on the larger ViT-L/14 backbone (43.3 vs. 42.9 mIoU).

% ******************  QKV attention

\begin{table}[!t]
  \tabcolsep5.0pt
  \centering
    % \scriptsize
    \tiny
  \begin{tabular}{c|c|lllllllll}
    \toprule
    CLIP &  Attention & V21 & C60 & Obj & V20 & C59 & Stf & City & ADE & Avg. \\
    \midrule
    % VIT b16 ****************************************
    \multirow{8}{*}{\rotatebox{90}{ViT-B/16}}  & Q-K & 34.8	& 19.0 & 21.4 & 76.1 & 21.5 & 14.2 & 15.9 & 10.7 & 26.7 \\
    & \cellcolor{gray!20.0}+ FSA & \cellcolor{gray!20.0}40.7 & \cellcolor{gray!20.0}22.2 & \cellcolor{gray!20.0}23.5 & \cellcolor{gray!20.0}76.9 & \cellcolor{gray!20.0}25.2 & \cellcolor{gray!20.0}16.4 & \cellcolor{gray!20.0}20.1 & \cellcolor{gray!20.0}12.3 & \cellcolor{gray!20.0}\bestimprovered{29.7}{3.0} \\
    \cmidrule(){2-11}
    &Q-Q & 55.6 & 31.7 & 32.9 & 81.4 & 35.5 & 23.9 & 30.3 & 17.9 & 38.7 \\
    & \cellcolor{gray!20.0}+ FSA  & \cellcolor{gray!20.0}56.9 & \cellcolor{gray!20.0}32.2 & \cellcolor{gray!20.0}34.2 & \cellcolor{gray!20.0}81.9 & \cellcolor{gray!20.0}35.9 & \cellcolor{gray!20.0}24.2 & \cellcolor{gray!20.0}31.2 & \cellcolor{gray!20.0}18.2 & \cellcolor{gray!20.0}\bestimprovered{39.3}{0.6} \\
    \cmidrule(){2-11}
    & K-K & 54.5 & 31.5 & 30.0 & 77.5 & 35.4 & 23.4 & 33.0 & 18.1 & 37.9 \\
    & \cellcolor{gray!20.0}+ FSA & \cellcolor{gray!20.0}57.5 & \cellcolor{gray!20.0}32.6 & \cellcolor{gray!20.0}33.3 & \cellcolor{gray!20.0}80.4 & \cellcolor{gray!20.0}36.2 & \cellcolor{gray!20.0}24.1 & \cellcolor{gray!20.0}33.5 & \cellcolor{gray!20.0}18.4 & \cellcolor{gray!20.0}\bestimprovered{39.5}{1.6} \\
    \cmidrule(){2-11}
    & V-V & 51.4 & 29.7 & 30.1 & 74.1 & 32.8 & 21.5 & 30.9 & 16.2 & 35.8 \\
    & \cellcolor{gray!20.0}+ FSA  & \cellcolor{gray!20.0}54.4 & \cellcolor{gray!20.0}31.4 & \cellcolor{gray!20.0}31.9 & \cellcolor{gray!20.0}78.4 & \cellcolor{gray!20.0}34.6 & \cellcolor{gray!20.0}22.6 & \cellcolor{gray!20.0}32.2 & \cellcolor{gray!20.0}17.2 & \cellcolor{gray!20.0}\bestimprovered{37.8}{2.0} \\
    \midrule
    % VIT - L14 **************************************
    \multirow{8}{*}{\rotatebox{90}{ViT-L/14}} & Q-K & 34.7 & 19.2 & 24.7 & 75.7 & 21.3 & 14.6 & 20.6 & 11.0 & 27.7 \\
    & \cellcolor{gray!20.0}+ FSA  & \cellcolor{gray!20.0}40.8 & \cellcolor{gray!20.0}22.0 & \cellcolor{gray!20.0}27.7 & \cellcolor{gray!20.0}77.3 & \cellcolor{gray!20.0}24.1 & \cellcolor{gray!20.0}16.5 & \cellcolor{gray!20.0}24.3 & \cellcolor{gray!20.0}12.2 & \cellcolor{gray!20.0}\bestimprovered{30.6}{2.9} \\
    \cmidrule(){2-11}
    & Q-Q & 49.2 & 26.7 & 31.1 & 80.1 & 29.5 & 19.8 & 30.6 & 16.6 & 35.5 \\
    & \cellcolor{gray!20.0}+ FSA  & \cellcolor{gray!20.0}50.4 & \cellcolor{gray!20.0}26.9 & \cellcolor{gray!20.0}31.4 & \cellcolor{gray!20.0}80.4 & \cellcolor{gray!20.0}29.8 & \cellcolor{gray!20.0}19.9 & \cellcolor{gray!20.0}30.7 & \cellcolor{gray!20.0}16.8 & \cellcolor{gray!20.0}\bestimprovered{35.8}{0.3} \\
    \cmidrule(){2-11}
    & K-K & 49.3 & 26.6 & 28.7 & 76.6 & 30.7 & 20.5 & 31.8 & 17.0 & 35.2 \\
    & \cellcolor{gray!20.0}+ FSA  & \cellcolor{gray!20.0}51.1 & \cellcolor{gray!20.0}27.3 & \cellcolor{gray!20.0}30.3 & \cellcolor{gray!20.0}78.5 & \cellcolor{gray!20.0}30.8 & \cellcolor{gray!20.0}20.4 & \cellcolor{gray!20.0}31.7 & \cellcolor{gray!20.0}17.2 & \cellcolor{gray!20.0}\bestimprovered{35.9}{0.7} \\
    \cmidrule(){2-11}
    & V-V & 47.4 & 27.0 & 31.0 & 78.7 & 29.8 & 20.2 & 31.1 & 18.0 & 35.4 \\
    & \cellcolor{gray!20.0}+ FSA  & \cellcolor{gray!20.0}49.4 & \cellcolor{gray!20.0}27.7 & \cellcolor{gray!20.0}31.4 & \cellcolor{gray!20.0}79.7 & \cellcolor{gray!20.0}30.6 & \cellcolor{gray!20.0}20.6 & \cellcolor{gray!20.0}31.8 & \cellcolor{gray!20.0}18.3 & \cellcolor{gray!20.0}\bestimprovered{36.2}{0.8} \\
    \midrule
    % VIT - H14 **************************************
    \multirow{8}{*}{\rotatebox{90}{ViT-H/14}}  & Q-K & 36.6 & 20.9 & 23.7 & 81.5 & 23.0 & 15.5 & 25.3 & 13.4 & 30.0 \\
    & \cellcolor{gray!20.0}+ FSA  & \cellcolor{gray!20.0}46.3 & \cellcolor{gray!20.0}24.7 & \cellcolor{gray!20.0}30.8 & \cellcolor{gray!20.0}83.5 & \cellcolor{gray!20.0}26.9 & \cellcolor{gray!20.0}18.1 & \cellcolor{gray!20.0}30.8 & \cellcolor{gray!20.0}15.5 & \cellcolor{gray!20.0}\bestimprovered{34.6}{4.6} \\
    \cmidrule(){2-11}
    & Q-Q & 51.1 & 26.8 & 29.7 & 79.1 & 29.9 & 19.6 & 32.7 & 17.7 & 35.8 \\
    & \cellcolor{gray!20.0}+ FSA  & \cellcolor{gray!20.0}52.8 & \cellcolor{gray!20.0}27.7 & \cellcolor{gray!20.0}30.5 & \cellcolor{gray!20.0}79.2 & \cellcolor{gray!20.0}30.8 & \cellcolor{gray!20.0}20.4 & \cellcolor{gray!20.0}34.0 & \cellcolor{gray!20.0}18.6 & \cellcolor{gray!20.0}\bestimprovered{36.7}{0.9} \\
    \cmidrule(){2-11}
    & K-K & 51.6 & 27.2 & 30.1 & 73.5 & 30.3 & 20.0 & 34.0 & 18.8 & 35.7 \\
    & \cellcolor{gray!20.0}+ FSA  & \cellcolor{gray!20.0}53.4 & \cellcolor{gray!20.0}28.1 & \cellcolor{gray!20.0}30.8 & \cellcolor{gray!20.0}75.0 & \cellcolor{gray!20.0}31.2 & \cellcolor{gray!20.0}20.8 & \cellcolor{gray!20.0}34.2 & \cellcolor{gray!20.0}19.5 & \cellcolor{gray!20.0}\bestimprovered{36.6}{0.9} \\
    \cmidrule(){2-11}
    & V-V & 51.2 & 28.1 & 30.0 & 80.2 & 31.0 & 21.3 & 33.7 & 19.7 & 36.9\\
    & \cellcolor{gray!20.0}+ FSA  & \cellcolor{gray!20.0}52.7 & \cellcolor{gray!20.0}28.6 & \cellcolor{gray!20.0}30.7 & \cellcolor{gray!20.0}80.5 & \cellcolor{gray!20.0}31.7 & \cellcolor{gray!20.0}21.6 & \cellcolor{gray!20.0}34.4 & \cellcolor{gray!20.0}19.9 & \cellcolor{gray!20.0}\bestimprovered{37.5}{0.6} \\
  \bottomrule
  \end{tabular}
    \caption{
   \textbf{Integration of our FSA into Q-K and self-self attentions.} Our method consistently shows improvements across all attention configurations. Specifically, we improve the original Q-K by a large margin, which is less effective in localization. 
}\vspace{-10pt}
    \label{tab:QKV_attn}
\end{table}

% \begin{tabular}{|c|c|c|}
% \hline
% \rotatebox{90}{Rotated Header} & Header 2 & Header 3 \\ \hline
% \rotatebox{90}{Text A}         & Data 1   & Data 2   \\ \hline
% \rotatebox{90}{Text B}         & Data 3   & Data 4   \\ \hline
% \end{tabular}
\begin{figure*}[h]
\centering

    %%%%%%%%%%%%%%%%%%%% Row 5 %%%%%%%%%%%%%%%%%%%%
    \begin{subfigure}{0.16\textwidth}
        \includegraphics[width=\linewidth]{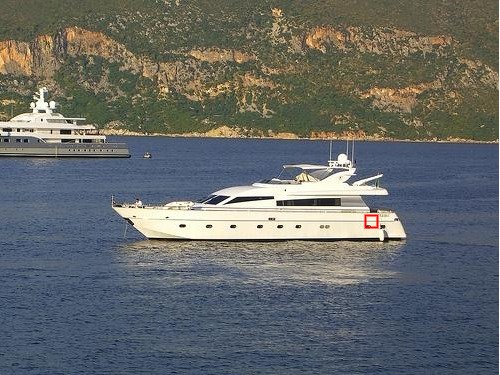}
    \end{subfigure}
    \begin{subfigure}{0.16\textwidth}
        \includegraphics[width=\linewidth]{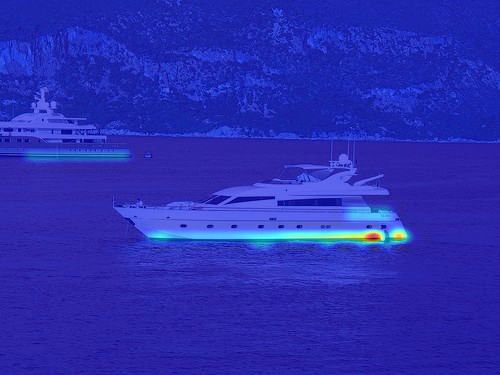}
    \end{subfigure}
    \begin{subfigure}{0.16\textwidth}
        \includegraphics[width=\linewidth]{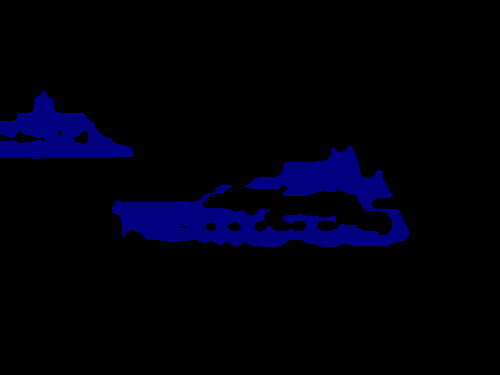}
    \end{subfigure}
    \begin{subfigure}{0.16\textwidth}
        \includegraphics[width=\linewidth]{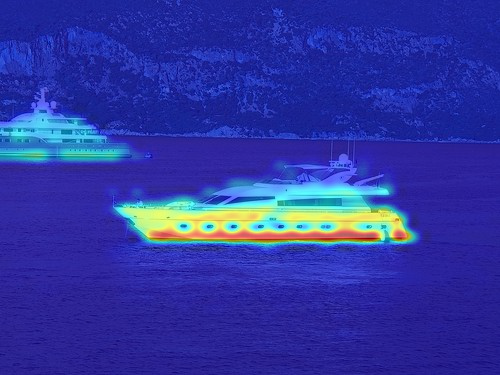}
    \end{subfigure}
    \begin{subfigure}{0.16\textwidth}
        \includegraphics[width=\linewidth]{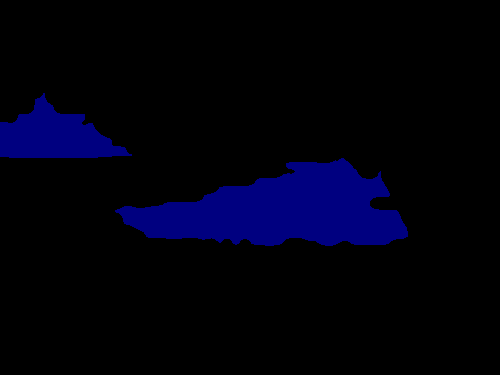}
    \end{subfigure}
    \begin{subfigure}{0.16\textwidth}
        \includegraphics[width=\linewidth]{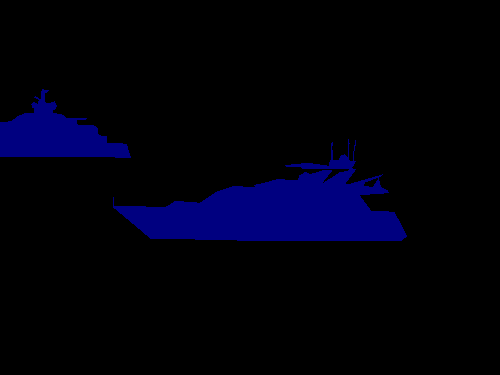}
    \end{subfigure}
    \\[1pt]

    %%%%%%%%%%%%%%%%%%%% Row 8 %%%%%%%%%%%%%%%%%%%%
    \begin{subfigure}{0.16\textwidth}
        \includegraphics[width=\linewidth]{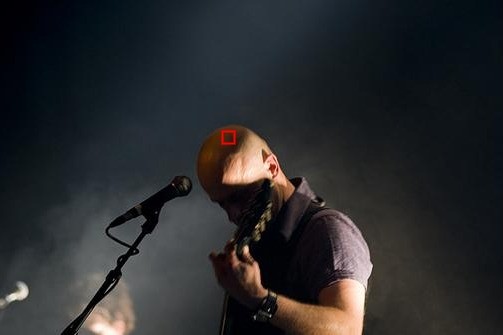}
        \caption*{\makecell{Input \\ image}}
    \end{subfigure}
    \begin{subfigure}{0.16\textwidth}
        \includegraphics[width=\linewidth]{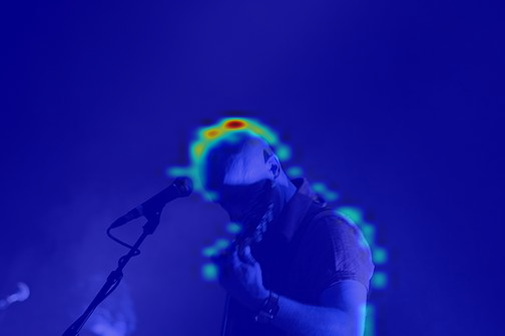}
        \caption*{\makecell{Attention map \\ of ProxyCLIP}}
    \end{subfigure}
    \begin{subfigure}{0.16\textwidth}
        \includegraphics[width=\linewidth]{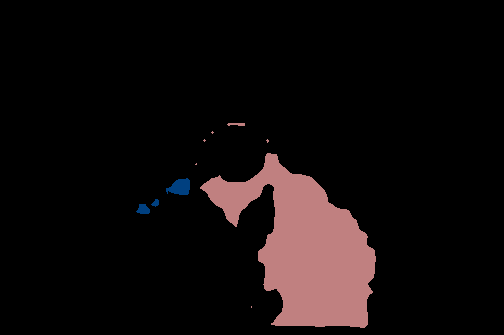}
        \caption*{\makecell{Segmentation map \\ of ProxyCLIP}}
    \end{subfigure}
    \begin{subfigure}{0.16\textwidth}
        \includegraphics[width=\linewidth]{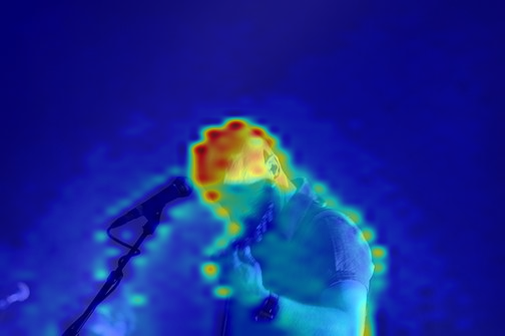}
         \caption*{\makecell{Our self-adapted \\ attention map}}
    \end{subfigure}
    \begin{subfigure}{0.16\textwidth}
        \includegraphics[width=\linewidth]{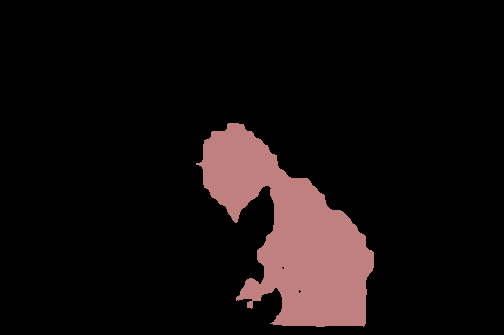}
        \caption*{\makecell{ProxyCLIP \\ + our FSA}}
    \end{subfigure}
    \begin{subfigure}{0.16\textwidth}
        \includegraphics[width=\linewidth]{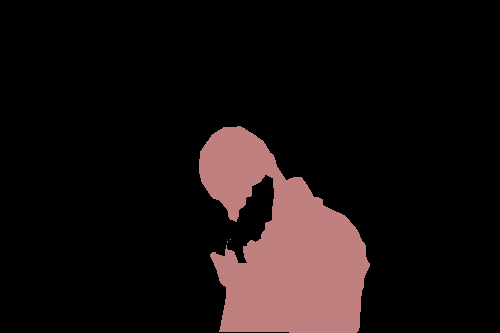}
        \caption*{\makecell{Ground truth \\ segmentation map}}
    \end{subfigure}
\caption{\textbf{Visualization of segmentation results of ProxyCLIP integrated with our FSA.} The attention maps ($2^{nd}$ and $4^{th}$ columns) correspond to the reference patch shown in the $1^{st}$ column. ProxyCLIP produces with holes within the same object due to weak attention across regions of the object. In contrast, our FSA effectively aggregates similar patches, enabling the correction of missegmented regions.
}
\vspace{-10pt}
\label{fig:qualitative_attn}
\end{figure*}

\begin{figure}[!ht]
    \centering
    
    % Row 1
    \begin{subfigure}{0.23\linewidth}
        \centering
        \includegraphics[width=\linewidth]{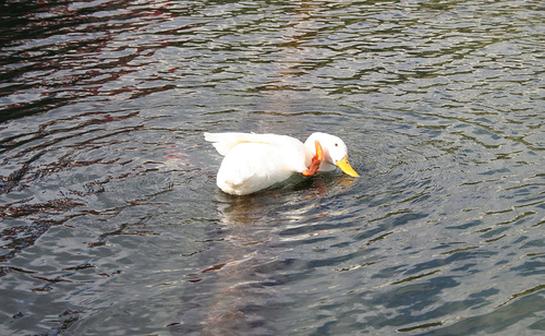 }
    \end{subfigure}
    \begin{subfigure}{0.23\linewidth}
        \centering
        \includegraphics[width=\linewidth]{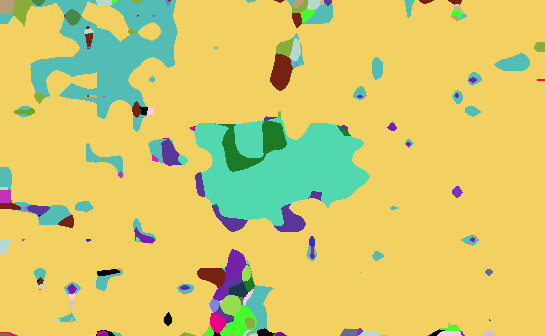}
    \end{subfigure}
    \begin{subfigure}{0.23\linewidth}
        \centering
        \includegraphics[width=\linewidth]{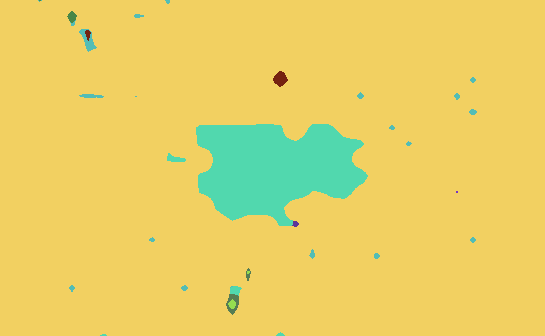}
    \end{subfigure}
    \begin{subfigure}{0.23\linewidth}
        \centering
        \includegraphics[width=\linewidth]{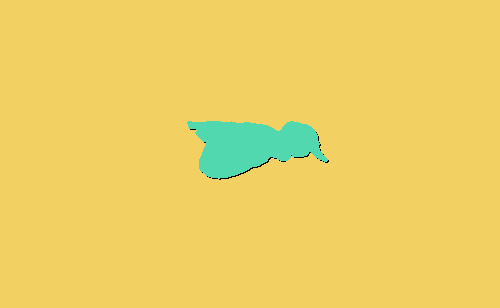}
    \end{subfigure}\\[1pt]
    % Row 2
    \begin{subfigure}{0.23\linewidth}
        \centering
        \includegraphics[width=\linewidth]{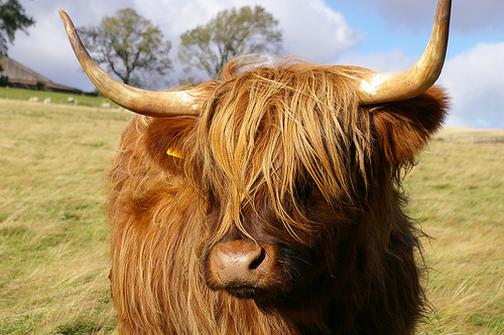 }
        \captionsetup{labelformat=empty}
        \caption{Input}
    \end{subfigure}
    \begin{subfigure}{0.23\linewidth}
        \centering
        \includegraphics[width=\linewidth]{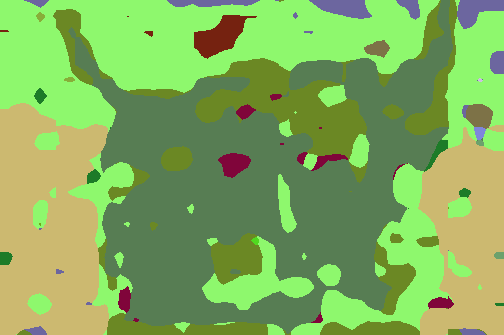}
        \captionsetup{labelformat=empty}
        \caption{MaskCLIP}
    \end{subfigure}
    \begin{subfigure}{0.23\linewidth}
        \centering
        \includegraphics[width=\linewidth]{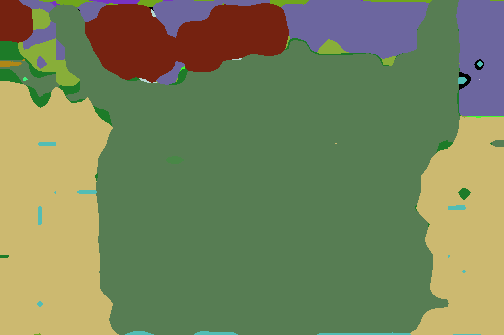}
        \captionsetup{labelformat=empty}
        \caption{+ FSA}
    \end{subfigure}
    \begin{subfigure}{0.23\linewidth}
        \centering
        \includegraphics[width=\linewidth]{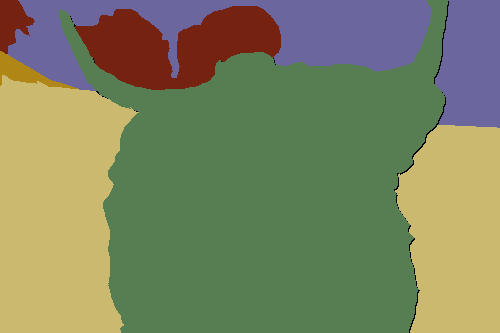}
        \captionsetup{labelformat=empty}
        \caption{GT}
    \end{subfigure}\\[1pt]

    % row 3
    \begin{subfigure}{0.23\linewidth}
        \centering
        \includegraphics[width=\linewidth]{ICCV_figs/qualitative/SCLIP/2007_001420/input.jpg }
    \end{subfigure}
    \begin{subfigure}{0.23\linewidth}
        \centering
        \includegraphics[width=\linewidth]{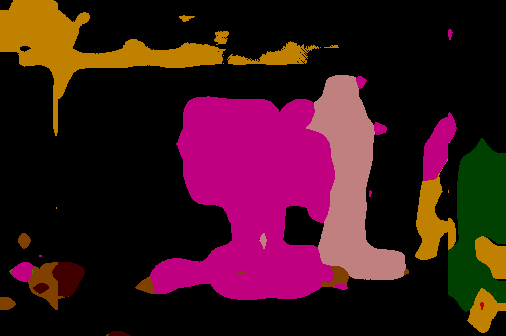}
    \end{subfigure}
    \begin{subfigure}{0.23\linewidth}
        \centering
        \includegraphics[width=\linewidth]{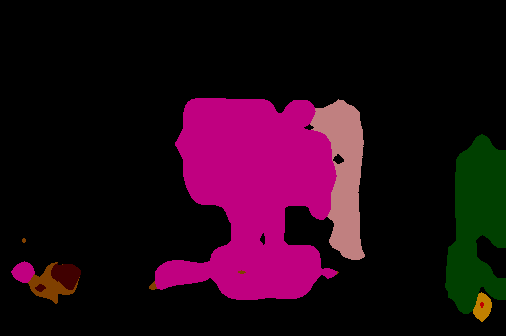}
    \end{subfigure}
    \begin{subfigure}{0.23\linewidth}
        \centering
        \includegraphics[width=\linewidth]{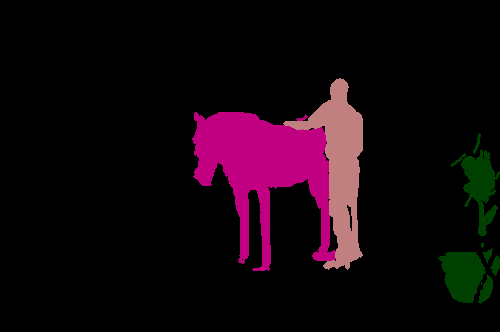}
    \end{subfigure}\\[1pt]
    % Row 4
    \begin{subfigure}{0.23\linewidth}
        \centering
        \includegraphics[width=\linewidth]{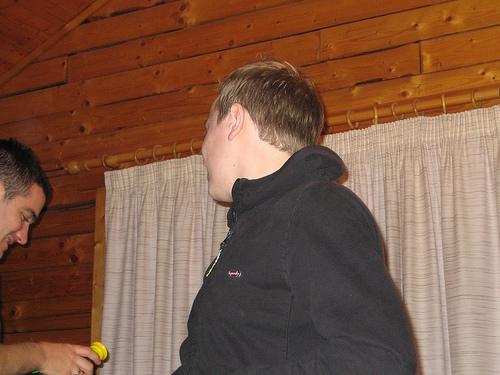 }
        \captionsetup{labelformat=empty}
        \caption{Input}
    \end{subfigure}
    \begin{subfigure}{0.23\linewidth}
        \centering
        \includegraphics[width=\linewidth]{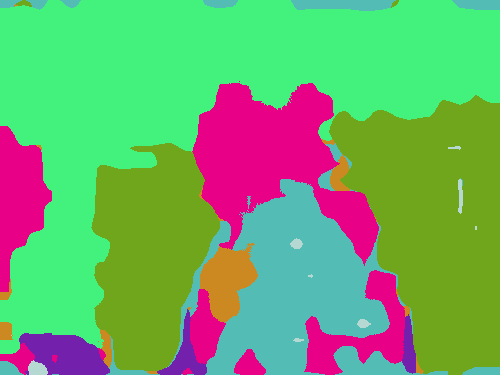}
        \captionsetup{labelformat=empty}
        \caption{SCLIP}
    \end{subfigure}
    \begin{subfigure}{0.23\linewidth}
        \centering
        \includegraphics[width=\linewidth]{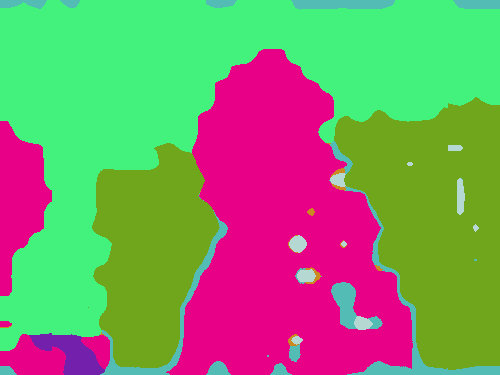}
        \captionsetup{labelformat=empty}
        \caption{+ FSA}
    \end{subfigure}
    \begin{subfigure}{0.23\linewidth}
        \centering
        \includegraphics[width=\linewidth]{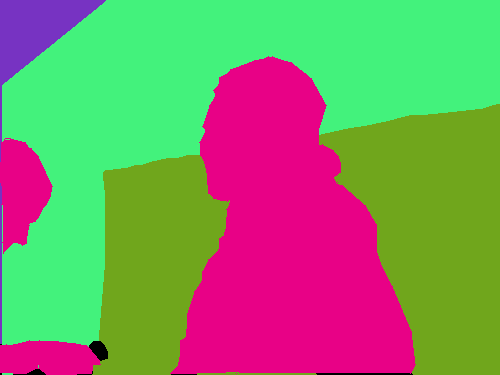}
        \captionsetup{labelformat=empty}
        \caption{GT}
    \end{subfigure}\\[1pt]
    
    % Row 5
    \begin{subfigure}{0.23\columnwidth}
        \centering
        \includegraphics[width=\linewidth]{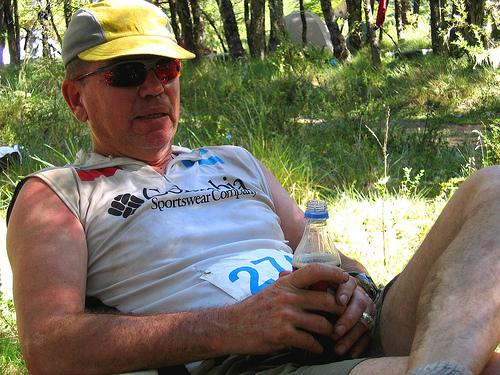}
    \end{subfigure}
    \begin{subfigure}{0.23\columnwidth}
        \centering
        \includegraphics[width=\linewidth]{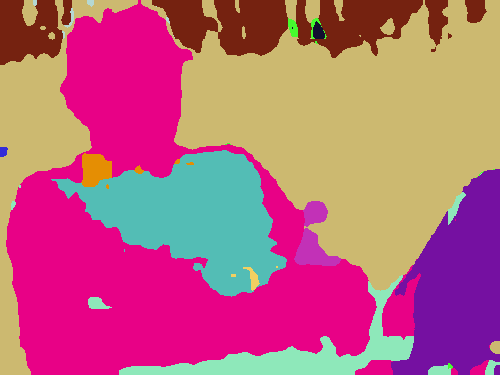}
    \end{subfigure}
    \begin{subfigure}{0.23\columnwidth}
        \centering
        \includegraphics[width=\linewidth]{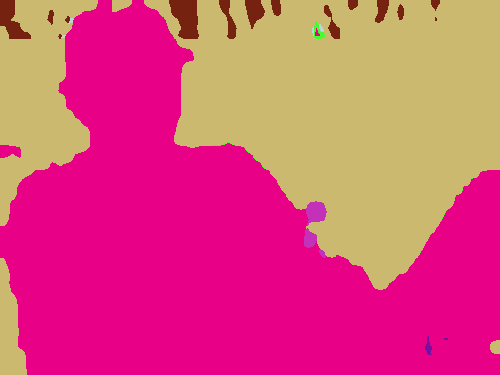}
    \end{subfigure}
    \begin{subfigure}{0.23\columnwidth}
        \centering
        \includegraphics[width=\linewidth]{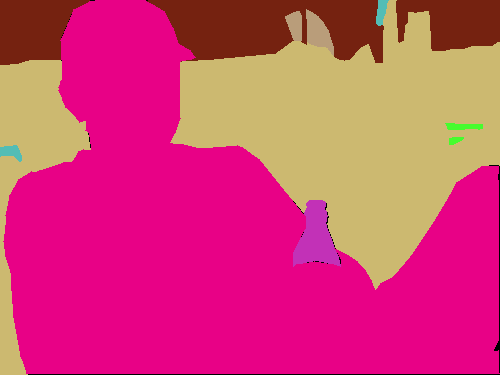}
    \end{subfigure}\\[1pt]
    
    % Row 6
    \begin{subfigure}{0.23\columnwidth}
        \centering
        \includegraphics[width=\linewidth]{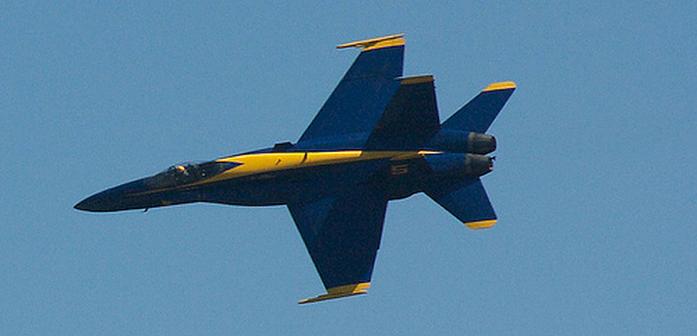}
        \captionsetup{labelformat=empty}
        \caption{Input}
    \end{subfigure}
    \begin{subfigure}{0.23\columnwidth}
        \centering
        \includegraphics[width=\linewidth]{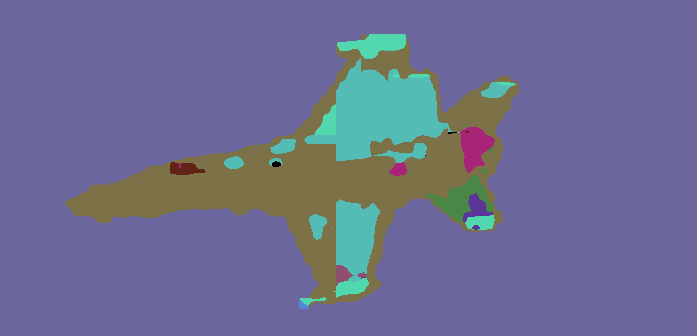}
        \captionsetup{labelformat=empty}
        \caption{ProxyCLIP}
    \end{subfigure}
    \begin{subfigure}{0.23\columnwidth}
        \centering
        \includegraphics[width=\linewidth]{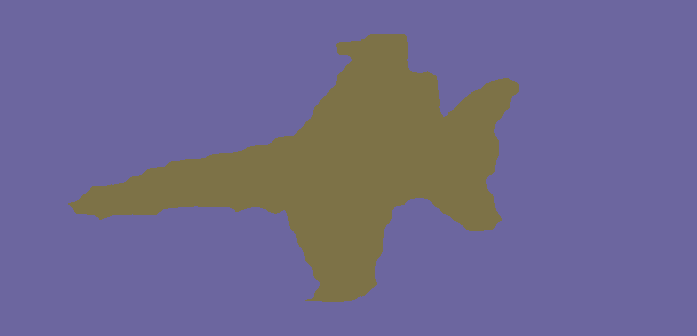}
        \captionsetup{labelformat=empty}
        \caption{+ FSA}
    \end{subfigure}
    \begin{subfigure}{0.23\columnwidth}
        \centering
        \includegraphics[width=\linewidth]{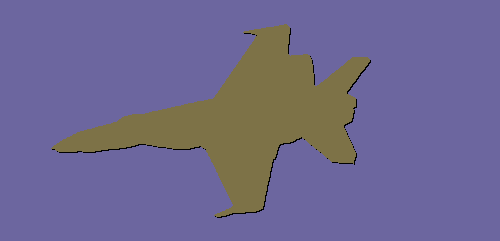}
        \captionsetup{labelformat=empty}
        \caption{GT}
    \end{subfigure}\\[2pt]
    \caption{\textbf{Qualitative results.} By integrating our feedback self-adaptive mechanism, we correct missegmented patches, ensuring consistent segmentation within each object.}
    \label{fig:qualitative}
    \vspace{-10pt}
\end{figure}

\noindent\textbf{Integration with various attention configurations.} 
The proposed feedback self-adaptive mechanism operates independently of the specific intermediate attention, demonstrating its versatility when integrated with various attention configurations. Table~\ref{tab:VFM_attn} shows results of applying our method to ProxyCLIP with different VFMs (MAE~\cite{he2022masked}, SAM~\cite{kirillov2023segment}, and DINOV2~\cite{oquab2023dinov2}), each displaying varying levels of spatial correspondence. Despite these differences, our method proves flexible, achieving consistent improvements across all VFMs, benchmarks, and backbones. Notably, while MAE shows weaker baseline performance, our feedback correction enhances it by \textbf{+1.2 mIoU}, \textbf{+2.5 mIoU}, and \textbf{+2.5 mIoU} across the three backbones.

We also integrate our method with inherent Q-K and self-self attention mechanisms using the codebase from ProxyCLIP. As summarized in Table~\ref{tab:QKV_attn}, improvements are observed across all attention types on the three backbones. The most significant gains are seen with Q-K attention (\textbf{+3.0 mIoU}, \textbf{+2.9 mIoU}, and \textbf{+4.6 mIoU}), as it is less effective for patch-level alignment. Overall, our self-adaptive framework is compatible with various attention configurations, yielding performance gains ranging from \textbf{+0.3 mIoU} to \textbf{+4.6 mIoU} as shown in Tables~\ref{tab:VFM_attn} and~\ref{tab:QKV_attn}.

\noindent\textbf{Qualitative comparison with SoTA.}
Fig.~\ref{fig:qualitative_attn} illustrates that ProxyCLIP often focuses on object edges, resulting in unstable segmentation. By integrating our adaptive FSA, we refine the intermediate attention to focus solely on the main regions of the same object, thereby improving the segmentation. Fig.~\ref{fig:qualitative} presents additional qualitative results with our FSA incorporated into MaskCLIP, SCLIP, and ProxyCLIP. Leveraging the feedback mechanism, we corrects misclassified regions, leading to more consistent segmentation. Additional results are provided in Appendix B.

\section{Ablation studies}
% In this section, we validate the effectiveness of different core components of our FSA.

\begin{table}[t]
\tabcolsep4pt
    \centering
    \small
    \begin{tabular}{cc|ccc}
        \toprule
             Scaling & Pruning & ViT-B/16 & ViT-L/14 & ViT-H/14 \\
        \midrule
        \ding{55} &\ding{55} & 22.6 & 23.5 & 25.5  \\
        \midrule
        \ding{55} &Fixed ratio& 36.6 & 37.8 & 41.0 \\
        \ding{55} &Fixed threshold& 37.2 & 38.6 & 40.1\\
        \ding{55} &Confidence-based& 38.2 & 39.4 & 42.5\\
        \checkmark &Confidence-based& 43.3 & 43.6 & 45.8\\
        \bottomrule
    \end{tabular}
    \caption{\textbf{Ablation on confidence-based pruning with ProxyCLIP over 8 benchmarks.} Our adaptive confidence-based pruning selects relevant patches while suppressing irrelevant ones. Additionally, adaptive exponential scaling emphasizes the importance of these selected patches, further enhancing their contribution.}
    \label{tab:ablation_pruning}
    \vspace{-10pt}
\end{table}
% \noindent\textbf{Confidence-based pruning.} 
% We generate sparse feedback attention using confidence-based pruning to highlight semantically related patches likely in the same class. As shown in Table~\ref{tab:ablation_pruning}, both fixed-ratio and threshold-based pruning improve segmentation by ignoring irrelevant patches, but confidence-based pruning yields greater gains through adaptive selection of correlated patches. Additionally, adaptive exponential scaling enhances relative distances between patch similarities, further emphasizing relevant patches and providing additional improvements.

\noindent\textbf{Confidence-based pruning.} 
We generate sparse feedback attention through confidence-based pruning, highlighting semantically related patches likely belonging to the same class. As shown in Table~\ref{tab:ablation_pruning}, both fixed-ratio and threshold-based pruning improve segmentation by filtering irrelevant patches, while confidence-based pruning achieves greater gains by adaptively selecting correlated patches. Additionally, adaptive exponential scaling amplifies relative differences in patch similarity, further enhancing segmentation.

\begin{figure}[t]
    \centering
    \includegraphics[width =\linewidth]{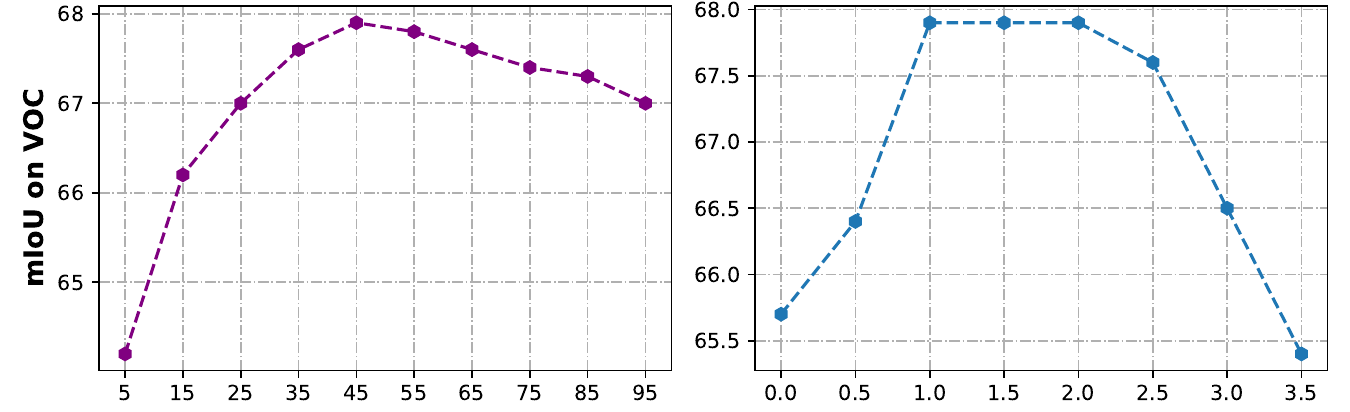} 
    \caption{
    % \textbf{Effect on \textit{p} (left) and $\lambda$ (right).} Increasing \textit{p} initially adds relevant patches, then irrelevant ones, causing performance to rise and then fall. $\lambda$ adjusts amplification and remains stable between 1 and 2.5.
    \textbf{Sensitivity on \textit{p} (left) and $\lambda$ (right) with ViT-L/14.} As \textit{p} increases, mIoU rises then falls with added irrelevant patches. $\lambda$ adjusts amplification and remains stable between 1 and 2.5.
}
    \label{fig:sensitivity}
    % \vspace{-2pt}
\end{figure}

Fig.~\ref{fig:sensitivity} illustrates the sensitivity of parameters \textit{p} and $\lambda$ using ViT-L/14 on the VOC dataset. As \textit{p} increases, more relevant patches are included; however, beyond a certain point, irrelevant patches are also incorporated, which affects attention on spatially correlated patches. Increasing $\lambda$ improves relative distances between patch similarities, allowing relevant patches to dominate. Performance remains stable between 1.0 and 2.5 but decreases when too few patches dominate, potentially excluding important patches. Fig.~\ref{fig:pruning} shows the distributions of row-wise pruning ratios, underscoring the necessity of an adaptive pruning mechanism.

\begin{table}[t]
    \centering
    \small
    \begin{tabular}{c|ccc}
        \toprule
         Attention isolation & MaskCLIP & SCLIP & ProxyCLIP \\
        \midrule
        w/o FSA (baseline) & 13.9 & 29.0 & 42.9 \\ 
        \midrule
        \ding{55} & \bestimprovered{29.8}{15.9} & \bestimprovered{33.2}{4.2} & \bestimprovered{43.4}{0.5}  \\
        \checkmark & \bestimprovered{\textbf{32.6}}{18.7} & \bestimprovered{\textbf{35.0}}{6.0} & \bestimprovered{\textbf{43.6}}{0.7} \\
        \bottomrule
    \end{tabular}
    \caption{\textbf{Validation on attention isolation with ViT-L/14.} Such isolation ensures that the output logits directly reflect the intermediate attention, enabling more consistent feedback adaptation.}
    \label{tab:ablation_uniform}
    \vspace{-10pt}
\end{table}
\begin{figure}[!ht]
    \centering
    
    % Row 1
    \begin{subfigure}{0.19\linewidth}
        \centering
        \includegraphics[width=\linewidth]{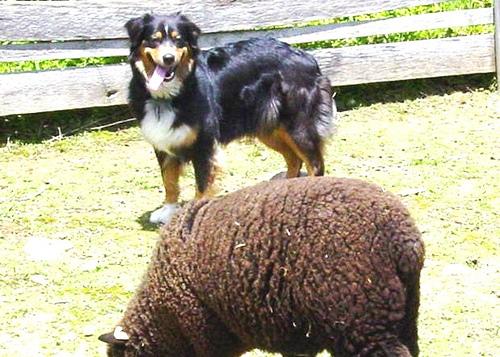}
    \end{subfigure}
    \begin{subfigure}{0.19\linewidth}
        \centering
        \includegraphics[width=\linewidth]{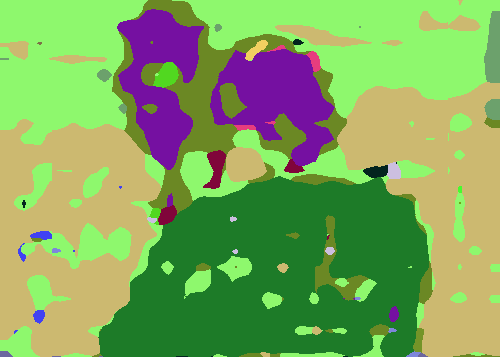}
    \end{subfigure}
    \begin{subfigure}{0.19\linewidth}
        \centering
        \includegraphics[width=\linewidth]{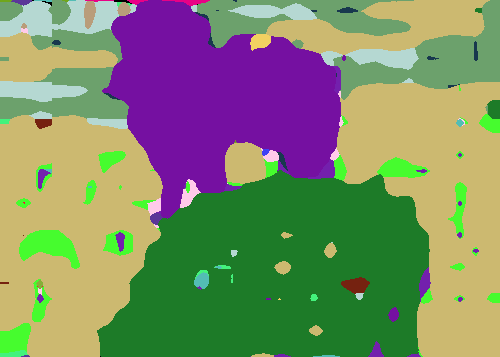}
    \end{subfigure}
    \begin{subfigure}{0.19\linewidth}
        \centering
        \includegraphics[width=\linewidth]{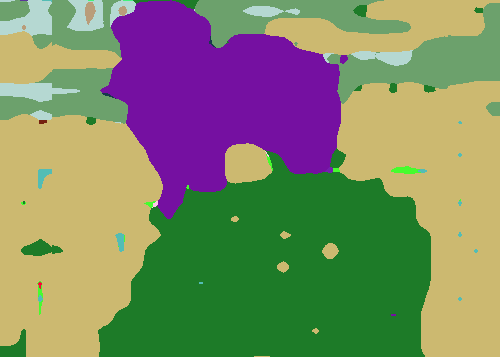}
    \end{subfigure}
    \begin{subfigure}{0.19\linewidth}
        \centering
        \includegraphics[width=\linewidth]{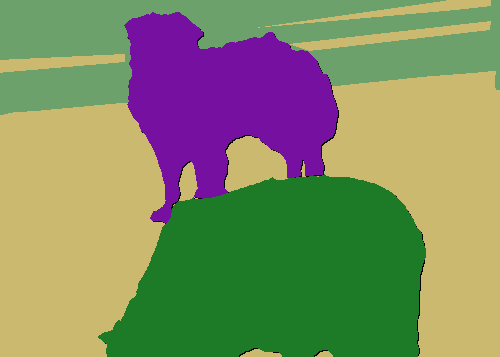}
    \end{subfigure}\\[1pt]
    
    % Row 2
    \begin{subfigure}{0.19\linewidth}
        \centering
        \includegraphics[width=\linewidth]{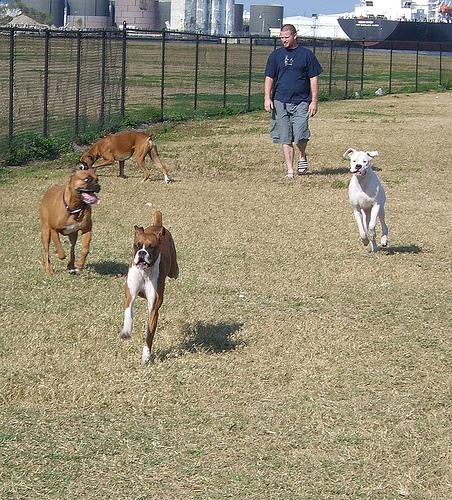 }
        \captionsetup{labelformat=empty}
        \caption{Input}
    \end{subfigure}
    \begin{subfigure}{0.19\linewidth}
        \centering
        \includegraphics[width=\linewidth]{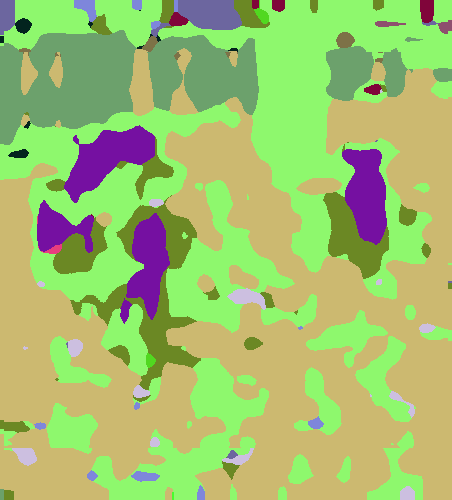}
        \captionsetup{labelformat=empty}
        \caption{MaskCLIP}
    \end{subfigure}
    \begin{subfigure}{0.19\linewidth}
        \centering
        \includegraphics[width=\linewidth]{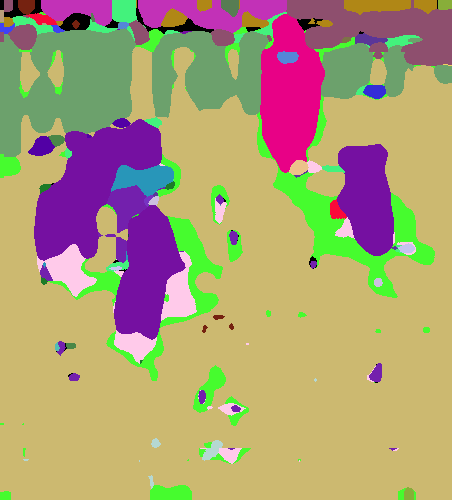}
        \captionsetup{labelformat=empty}
        \caption{FSA w/o iso.}
    \end{subfigure}
    \begin{subfigure}{0.19\linewidth}
        \centering
        \includegraphics[width=\linewidth]{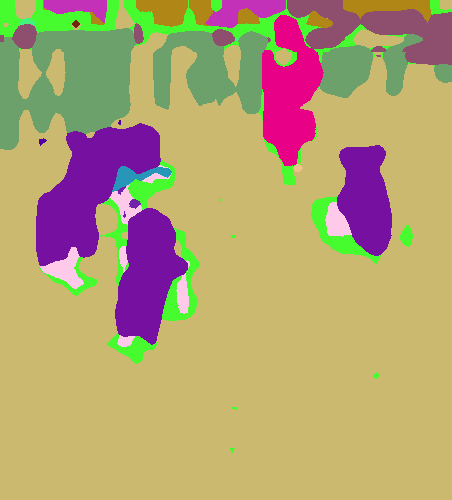}
        \captionsetup{labelformat=empty}
        \caption{FSA w/ iso.}
    \end{subfigure}
    \begin{subfigure}{0.19\linewidth}
        \centering
        \includegraphics[width=\linewidth]{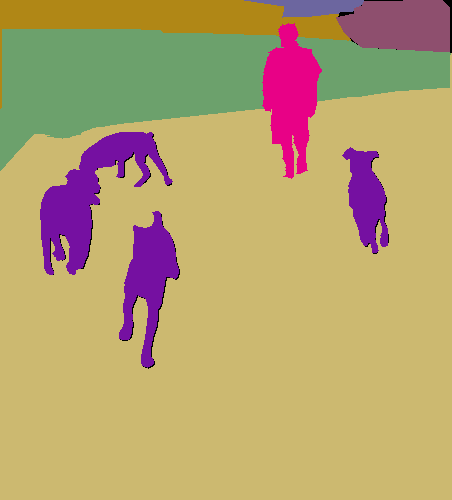}
        \captionsetup{labelformat=empty}
        \caption{GT}
    \end{subfigure}\\[1pt]
    \caption{\textbf{Qualitative results on attention isolation (iso.)}.}
    \label{fig:qualitative_debias}
    \vspace{-10pt}
\end{figure}

\noindent\textbf{Attention isolation.}  
Our feedback attention modulates the intermediate attention, making it crucial that output relationships reflect only the initial attention maps. As shown in Table~\ref{tab:ablation_uniform}, we observe improvements over 3 methods even without isolation. When attention isolation is applied, the output logits align better with the intermediate attention, leading to further segmentation gains. The qualitative examples in Fig.~\ref{fig:qualitative_debias} also demonstrate that attention isolation is effective in removing noise in segmentation maps.

\noindent\textbf{Ensembling on adapted attentions.} 
Table~\ref{tab:ablation_ensemble} presents the individual segmentation results for each adaptation strategy in Eqs.~\ref{eq:1}-\ref{eq:3}. While each method shows improvement, the best performance varies across SoTA methods. Our ensembling strategy combines their strengths, resulting in consistent and enhanced segmentation performance.

\begin{figure}[t]
    \centering
    \includegraphics[width =\linewidth]{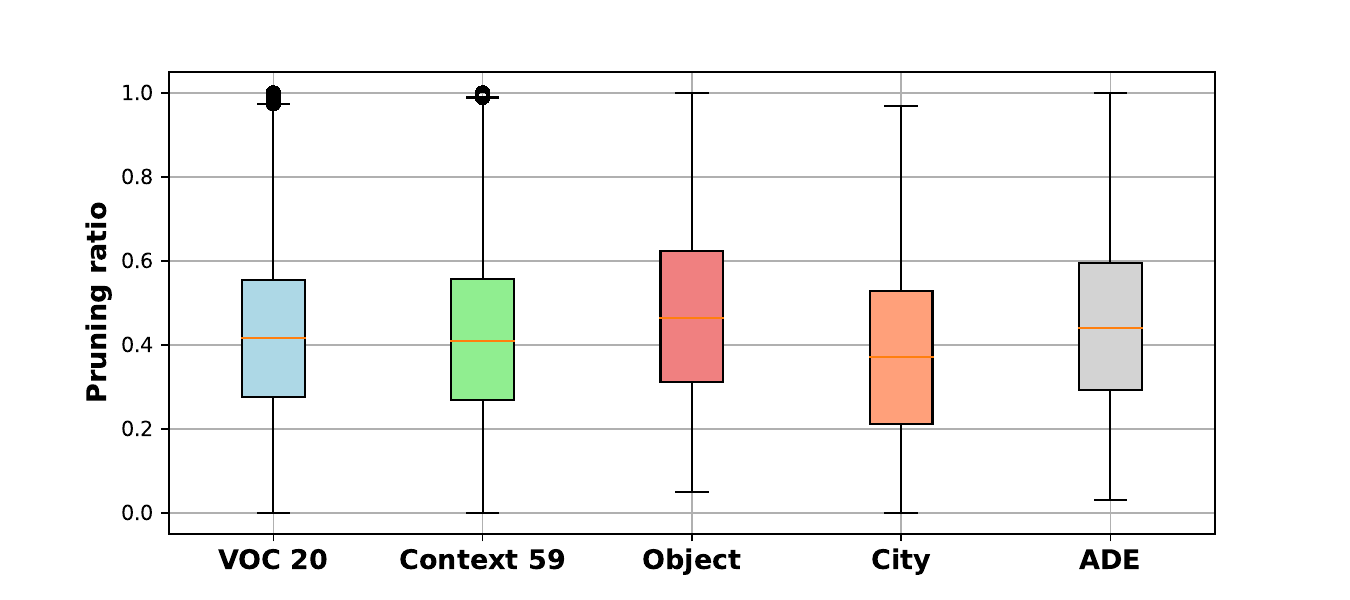} 
    \caption{\textbf{Distribution of pruning ratio.} The variety of distribution demonstrates the necessity of our adaptive mechanism. }
    \label{fig:pruning}
    % \vspace{-2pt}
\end{figure}

\begin{table}[t]
\tabcolsep3pt
    \centering
    \small
    \begin{tabular}{ccc|ccc}
        \toprule
             $\boldsymbol{Y}^{dense}_1$ & $\boldsymbol{Y}^{dense}_2$ & $\boldsymbol{Y}^{dense}_3$ & MaskCLIP & SCLIP & ProxyCLIP \\
         \midrule
         \multicolumn{3}{c|}{w/o FSA (baseline)} & 13.9 & 29.0 & 42.9 \\
         \midrule
        \checkmark &&& \bestimprovered{32.1}{18.2} & \bestimprovered{34.2}{5.2} & \bestimprovered{43.5}{0.6} \\
         &\checkmark&& \bestimprovered{31.8}{17.9} & \bestimprovered{33.9}{4.9} & \bestimprovered{43.4}{0.5} \\
         &&\checkmark& \bestimprovered{30.9}{17.0} & \bestimprovered{34.5}{5.5} & \bestimprovered{43.5}{0.6} \\
        \multicolumn{3}{c|}{Ensenble}& \bestimprovered{\textbf{32.6}}{18.7} & \bestimprovered{\textbf{35.0}}{6.0} & \bestimprovered{\textbf{43.6}}{0.7} \\
        \bottomrule
    \end{tabular}
    \caption{\textbf{Validation on ensembling with ViT-L/14.} Our ensembling strategy combines the strengths of each adaptation, resulting in consistent and enhanced segmentation across methods.}
    \label{tab:ablation_ensemble}
    % \vspace{-10pt}
\end{table}

\begin{table}
    \centering
    % \tiny
    \small
    \tabcolsep3pt
    \begin{tabular}{c|ccc}
        \toprule
        Methods & ViT-B/16 & ViT-L/14 & ViT-H/14 \\
        \midrule
        Proxy & 12.9ms & 20.1ms & 28.1ms \\
        + FSA & 13.6ms & 21.1ms & 29.2ms \\
        \bottomrule
    \end{tabular}
    \caption{\textbf{Speed on V100 GPU with patch resolution of 224x224.}} % Example caption, make sure to complete it
    \label{tab:cost}
    \vspace{-5pt}
\end{table}

\noindent\textbf{Cost analysis.}  Table~\ref{tab:cost} reports speed averaged over 3 trails of 100 forward passes. As we only modify the last attention as in ProxyCLIP, FSA only adds 3-5\% overhead.

\noindent\textbf{Iterative adaptation.} Intuitively, our FSA supports iterative adaptation, but, further improvement is not observed.

\section{Conclusion}
% In conclusion, we introduce FSA, a training-free feedback self-adaptive attention mechanism for open-vocabulary semantic segmentation. By enhancing intermediate attention through feedback based on patch logits correlations, our method ensures spatial consistency and incorporates class information, addressing the coherence inconsistencies of prior approaches. With modules like bias removal, confidence-based sparse pruning, and adaptation ensemble, our framework demonstrates effectiveness across 8 benchmarks with 24 attention configurations.

% In this work, we introduce FSA, a novel training-free self-adaptive framework that feedback on the spatial coherence of attention maps by leveraging the patch prediction from the model itself. Our method effectively bridges the gap between intermediate attention and final predictions by integrating both visual and textual cues, enabling more accurate patch-level correspondence and boosting overall segmentation accuracy. On 8 standard benchmarks, our FSA consistently improves over 24 attention configurations and achieves the SoTA performance with our best model.

In this work, we introduce FSA, a novel training-free self-adaptive framework that enhances spatial coherence in attention maps by leveraging the model’s own patch predictions as feedback. Our method effectively bridges the gap between intermediate attention and final outputs by integrating visual and textual cues, resulting in more accurate patch-level correspondence and improved segmentation. Integrated into various attention configurations, FSA consistently improves on 8 standard benchmarks.

% If the network ultimately decides these patches belong to the same class, it should also reflect that in the attention structure—so let’s feed that information back into the attention computation.

% There is often no direct mechanism to ensure those earlier improvements carry over to the final predictions.

{
    \small
    \bibliographystyle{ieeenat_fullname}
    \bibliography{main}
}
\newpage

\setlength{\abovedisplayskip}{5pt}
\setlength{\belowdisplayskip}{5pt}
\setlength{\belowcaptionskip}{-8pt}

\setcounter{section}{0}
% \makeatletter
\renewcommand{\theequation}{S\arabic{equation}}
\renewcommand{\thefigure}{S\arabic{figure}}
\renewcommand{\thetable}{S\arabic{table}}
\renewcommand{\thesection}{\Alph{section}}

\section{Summary}
In this supplementary material, we present the following additional content to complement the main paper:
\begin{itemize}
    \item Additional qualitative comparisons on various datasets.
    \item We present more details motivation and additional observations.
    \item Sensitivity on similarity metric.
    \item Impact of different configuration of CLIP.
    \item Additional speed analysis.
\end{itemize}

\section{Additional qualitative results}
\begin{figure}[!ht]
    \centering
    
    % Row 1
    \begin{subfigure}{0.23\columnwidth}
        \centering
        \includegraphics[width=\linewidth]{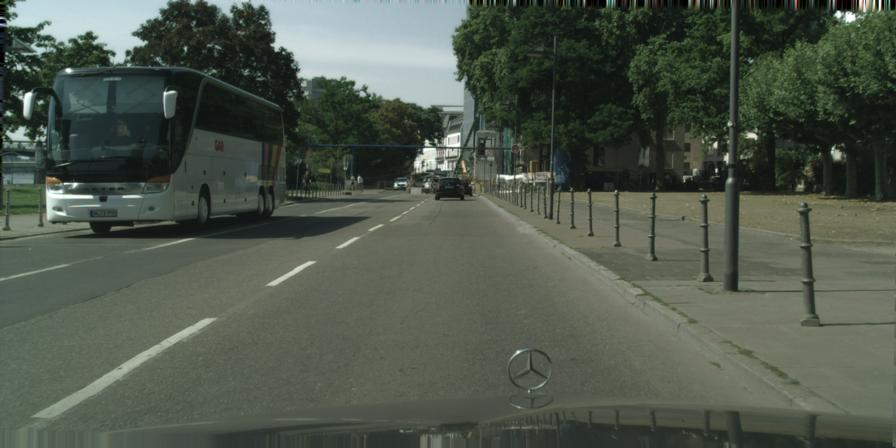}
    \end{subfigure}%
    \begin{subfigure}{0.23\columnwidth}
        \centering
        \includegraphics[width=\linewidth]{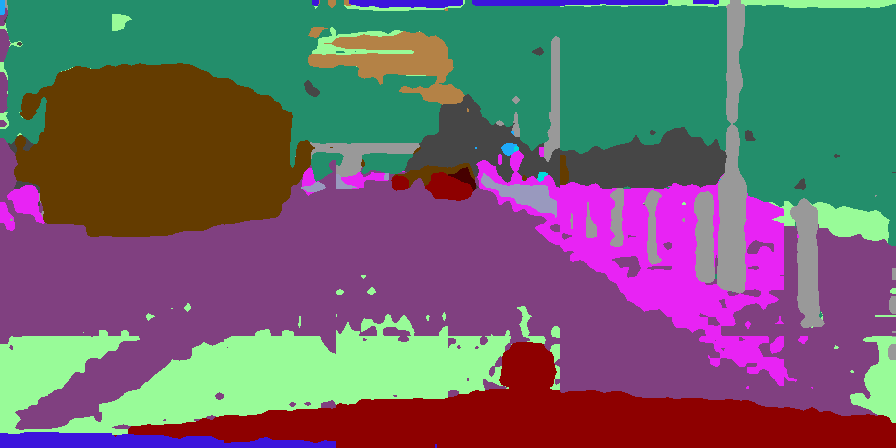}
    \end{subfigure}%
    \begin{subfigure}{0.23\columnwidth}
        \centering
        \includegraphics[width=\linewidth]{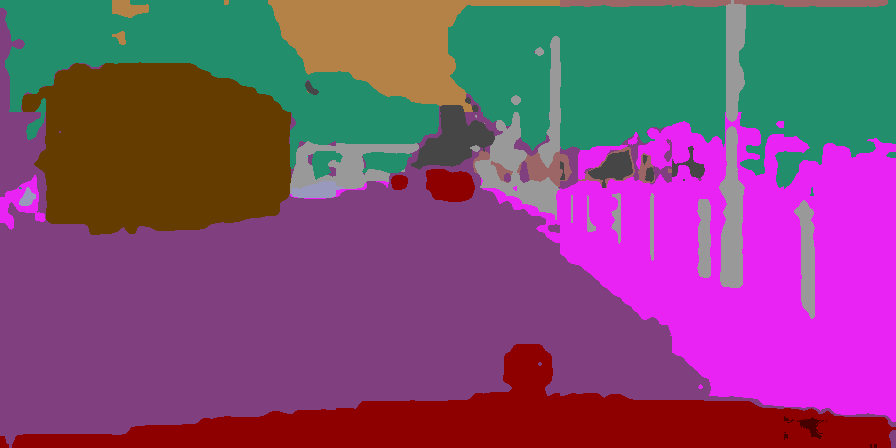}
    \end{subfigure}%
    \begin{subfigure}{0.23\columnwidth}
        \centering
        \includegraphics[width=\linewidth]{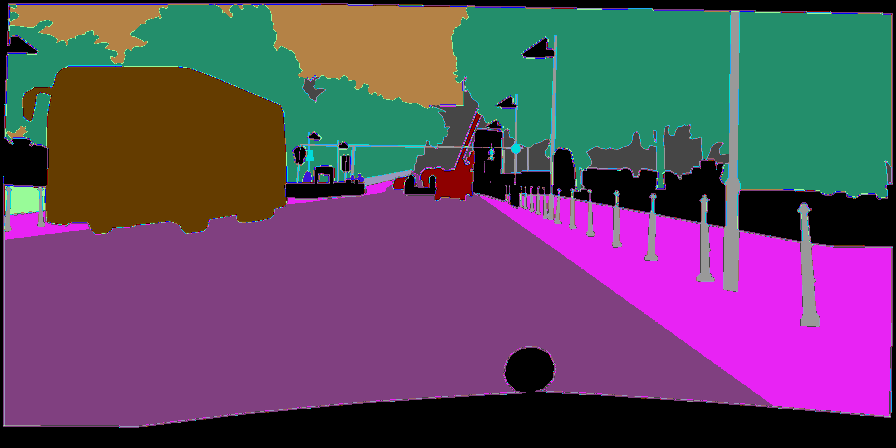}
    \end{subfigure}\\[2pt]
    % Row 2
    \begin{subfigure}{0.23\columnwidth}
        \centering
        \includegraphics[width=\linewidth]{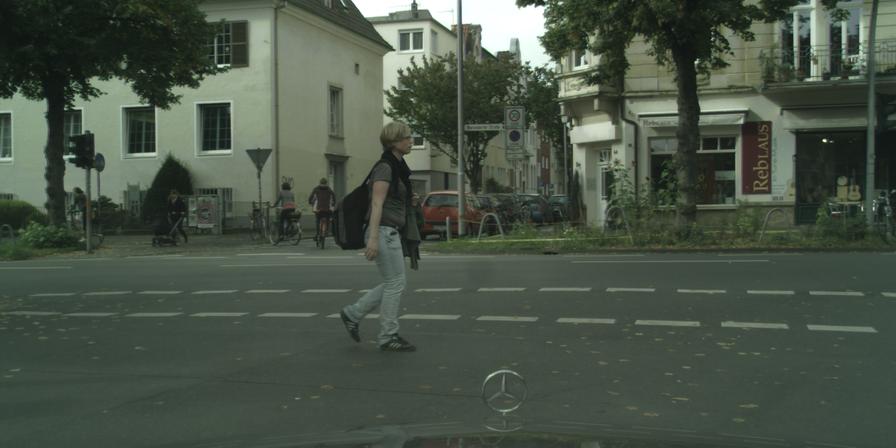}
    \end{subfigure}%
    \begin{subfigure}{0.23\columnwidth}
        \centering
        \includegraphics[width=\linewidth]{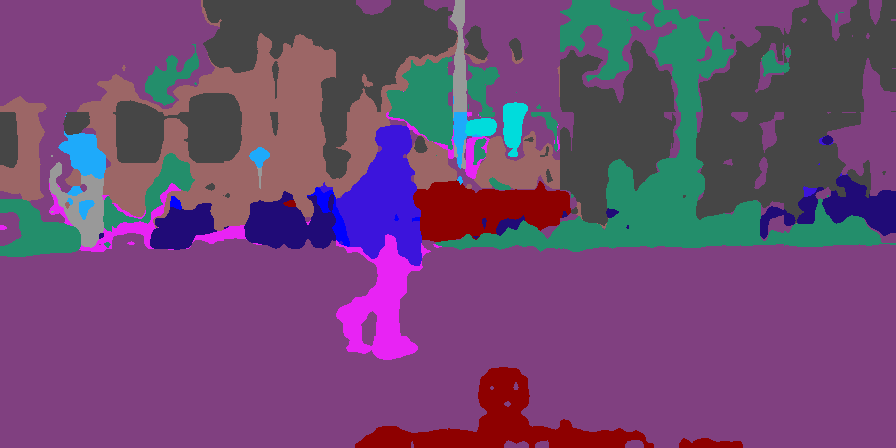}
    \end{subfigure}%
    \begin{subfigure}{0.23\columnwidth}
        \centering
        \includegraphics[width=\linewidth]{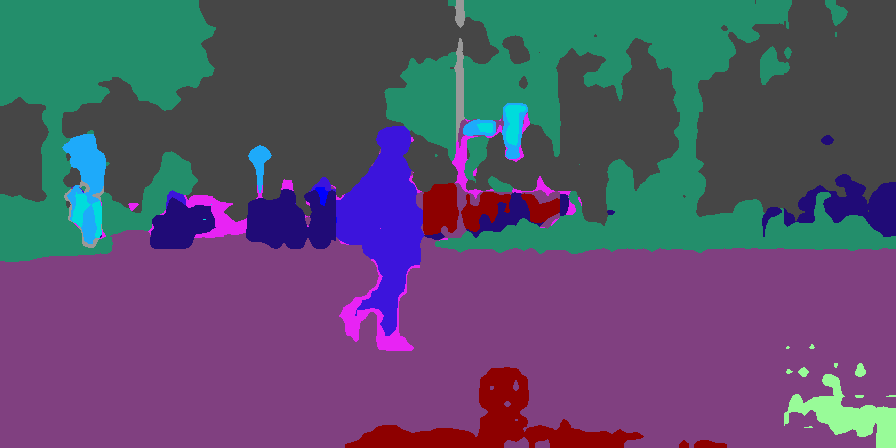}
    \end{subfigure}%
    \begin{subfigure}{0.23\columnwidth}
        \centering
        \includegraphics[width=\linewidth]{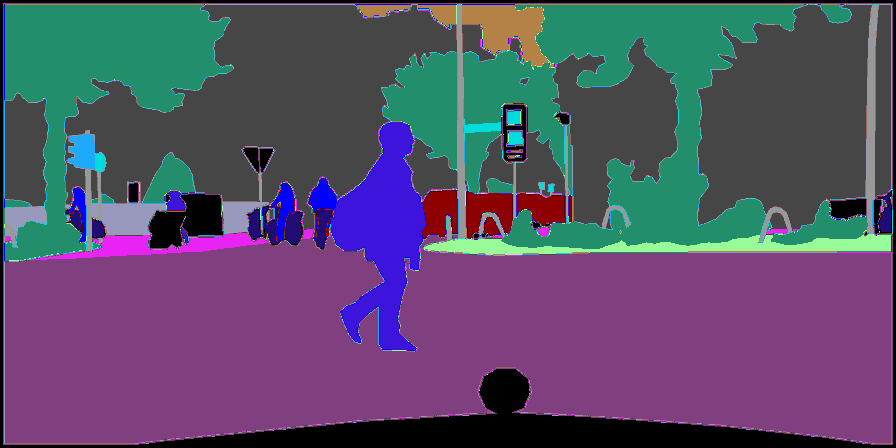}
    \end{subfigure}\\[2pt]
    
    % Row 3
    \begin{subfigure}{0.23\columnwidth}
        \centering
        \includegraphics[width=\linewidth]{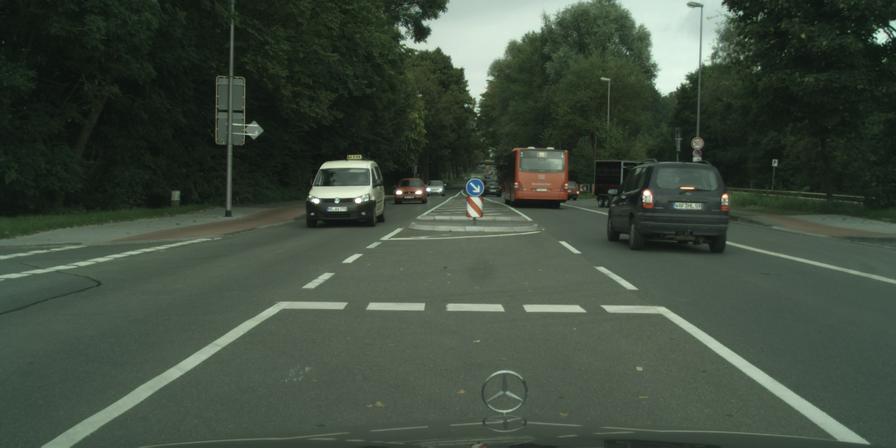}
    \end{subfigure}%
    \begin{subfigure}{0.23\columnwidth}
        \centering
        \includegraphics[width=\linewidth]{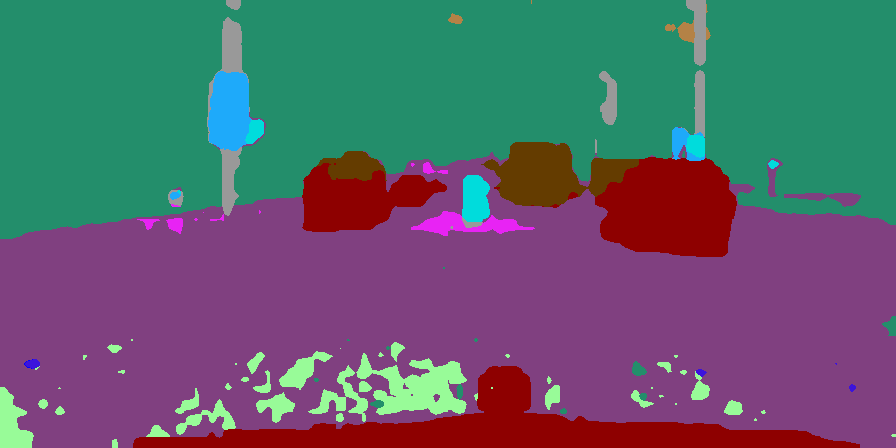}
    \end{subfigure}%
    \begin{subfigure}{0.23\columnwidth}
        \centering
        \includegraphics[width=\linewidth]{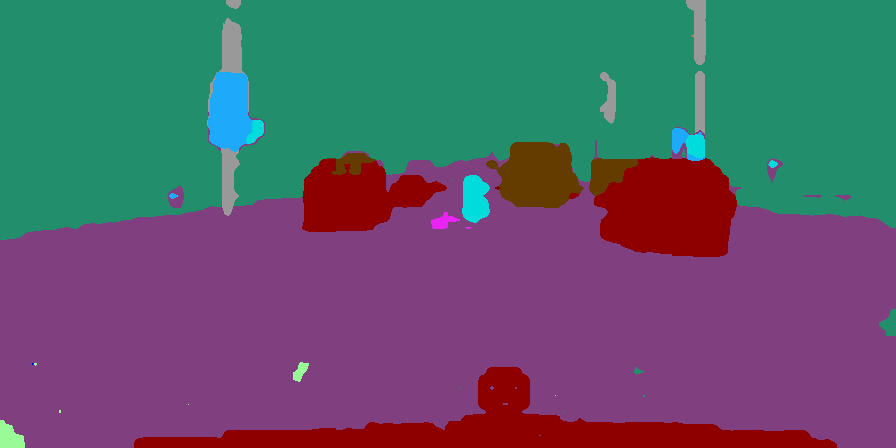}
    \end{subfigure}%
    \begin{subfigure}{0.23\columnwidth}
        \centering
        \includegraphics[width=\linewidth]{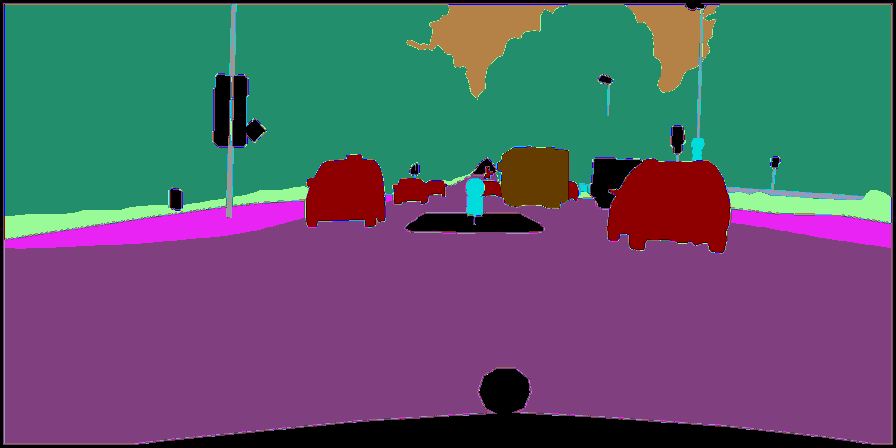}
    \end{subfigure}\\[2pt]
    
    % Row 4
    \begin{subfigure}{0.23\columnwidth}
        \centering
        \includegraphics[width=\linewidth]{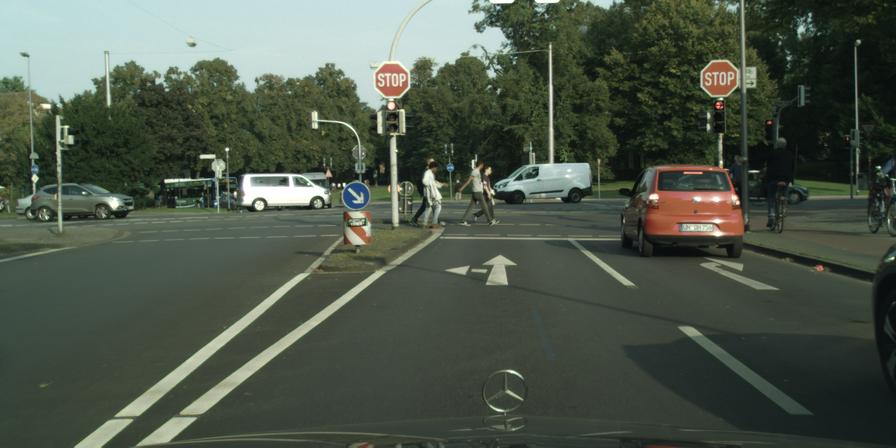}
        \caption{Input}
    \end{subfigure}%
    \begin{subfigure}{0.23\columnwidth}
        \centering
        \includegraphics[width=\linewidth]{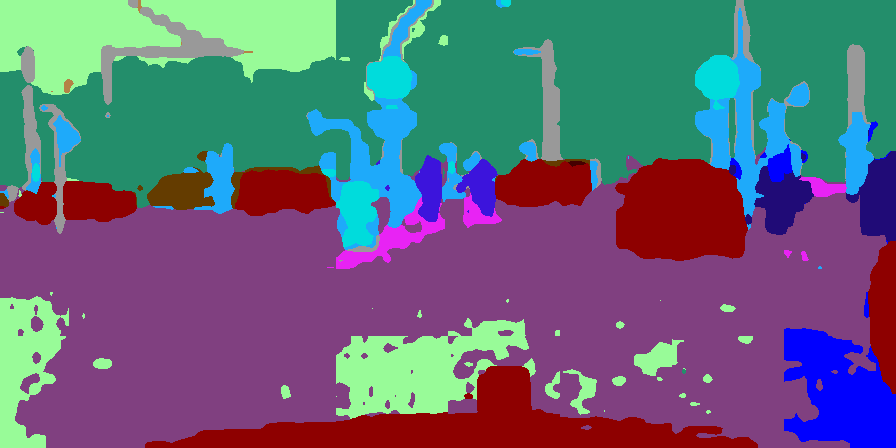}
        \caption{ProxyCLIP}
    \end{subfigure}%
    \begin{subfigure}{0.23\columnwidth}
        \centering
        \includegraphics[width=\linewidth]{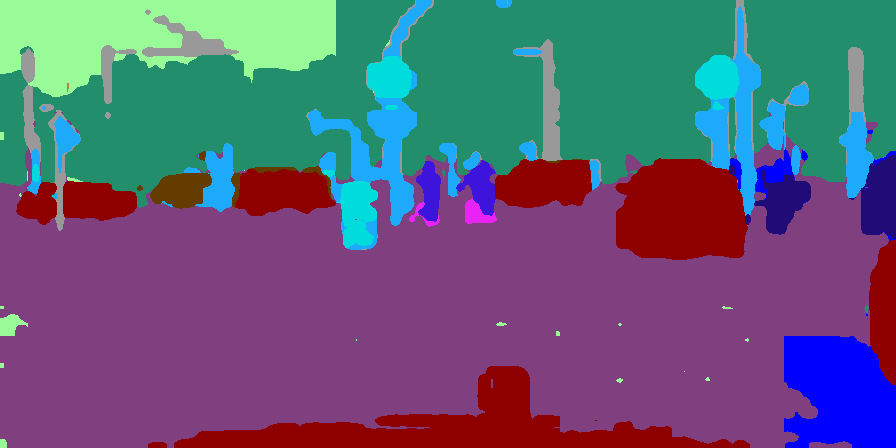}
        \caption{+FSA}
    \end{subfigure}%
    \begin{subfigure}{0.23\columnwidth}
        \centering
        \includegraphics[width=\linewidth]{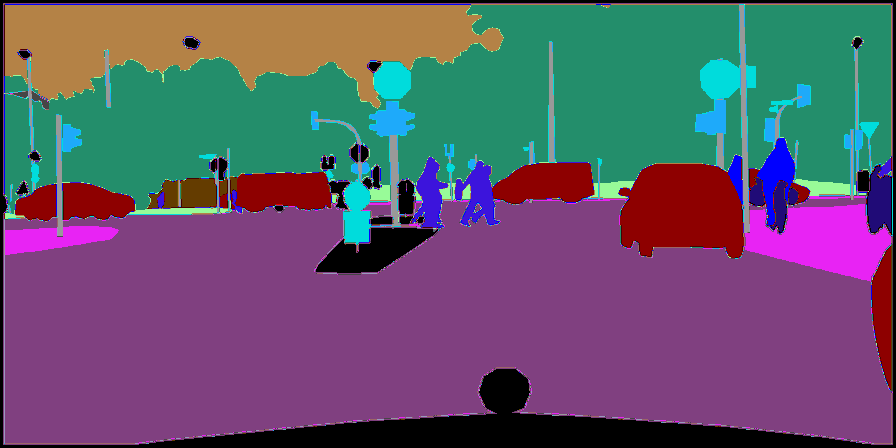}
        \caption{GT}
    \end{subfigure}\\[2pt]
    
    \caption{\textbf{Qualitative results on Cityscapes.} By integrating our feedback self-adaptive mechanism, we correct missegmented patches by ProxyCLIP, ensuring consistent segmentation within each object. We can clearly observe that our segmentation is more consistent across whole objects.}
    \label{fig:qualitative_city}
\end{figure}

\begin{figure}[h]
    \centering
    % \vspace{-10pt}
    \includegraphics[width =1.0\linewidth]{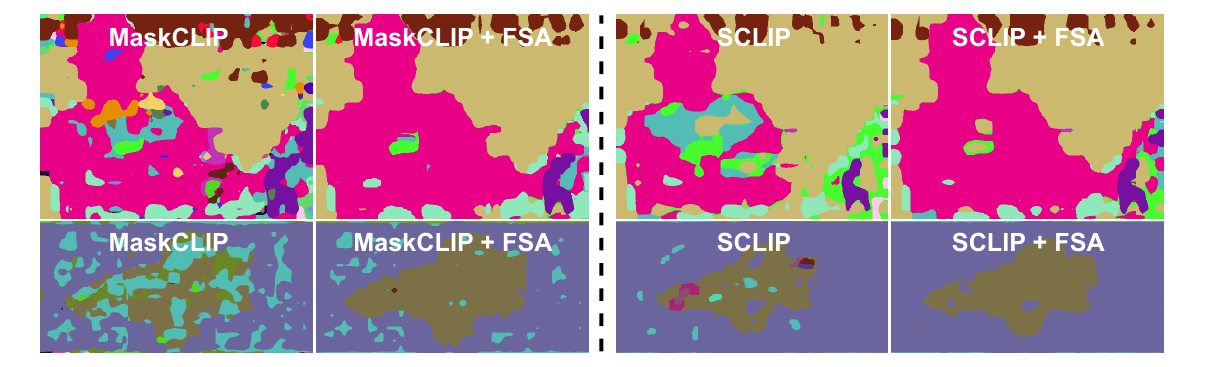}
    \caption{\textbf{Examples of Row 5-6 in Fig.6 for MaskCLIP and SCLIP.}}
    \label{fig:compare}
\end{figure}

In Fig.~\ref{fig:qualitative_city}, we provide additional qualitative comparisons with ProxyCLIP on the Cityscapes dataset. By incorporating our self-adaptive framework, we successfully correct missegmented regions. Notably, for the same object, certain regions are initially misclassified; however, our feedback-adaptive method aggregates information from similar patches in the output, enabling further refinement and correction. In Fig.~\ref{fig:compare}, we show expanded comparison with MaskCLIP and SCLIP with the exmaples in Row 5-6 of Fig. 6.

In Fig.~\ref{fig:supp_voc}, we present additional results from the VOC21 dataset, along with attention maps corresponding to the reference patch (indicated by the red box in the first column). The segmentation results of ProxyCLIP (third column) exhibit flaws, as certain regions within the main object are incorrectly segmented. This issue arises because those patches fail to attend correctly to the same object, as illustrated in their attention maps (second column). In contrast, our feedback self-adaptive method successfully corrects the segmentation (fifth column) across the entire object by attending to more regions belonging to the same object.

\section{Additional motivation and observation}

Our proposed FSA aims to improve the spatial coherence among similar patches using the feedback loop. The feedback loop is derived using self-predicted logits for each patch. The concept is similar to knowledge distillation~\cite{zhong2022meta,hinton2015distilling}, where the output logits of a stronger model is used as a soft label to guide the current model to learn extended knowledge, instead of the sparse labels from ground truth. More specifically, it is close to self-distillation~\cite{zhang2021self,zhang2019your} where both the teacher and student are the model itself. On the other hand, our methodology is also closely related to test-time adaptation, which normally adapt the model towards one specific test data instance~\cite{wang2024distribution,wang2020tent} or specific domain~\cite{wu2024test,chi2025learning}. The process is normally self-supervised without any additional manual labeling~\cite{liu2021ttt++,shu2022test}.

\begin{figure}[h]
    \centering
    % \vspace{-10pt}
    \includegraphics[width =0.9\linewidth]{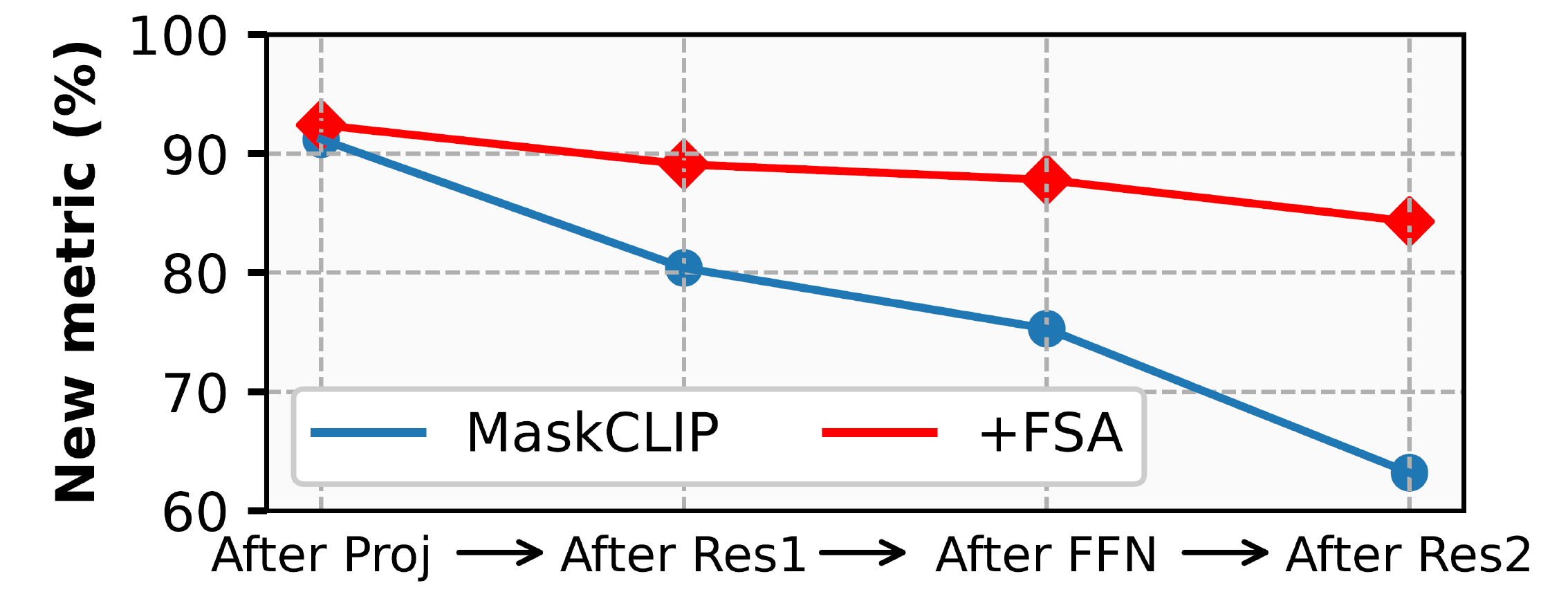}
    \includegraphics[width =0.9\linewidth]{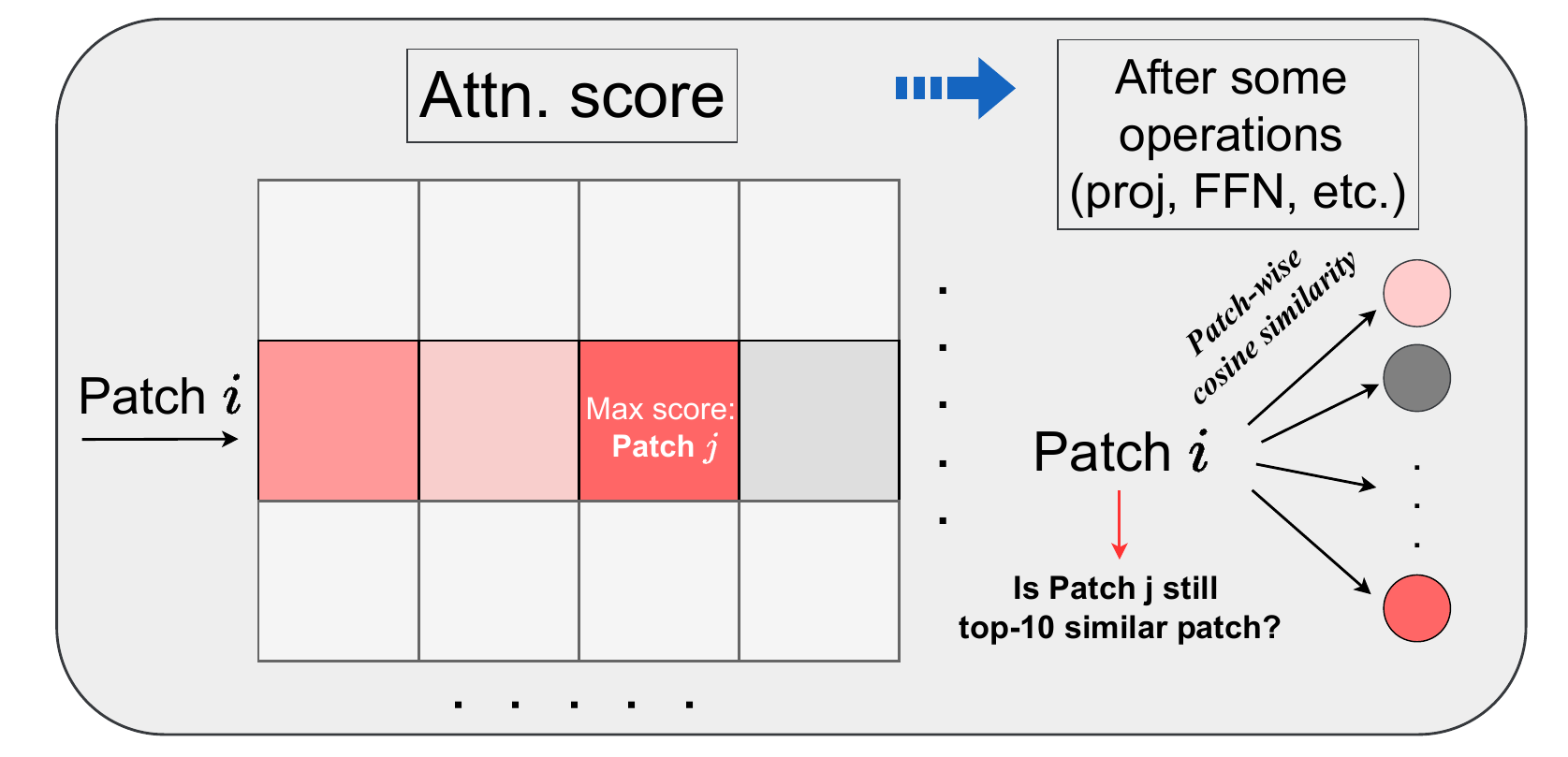}
    \caption{Illustration and analysis for new metric.}
    \label{fig:new_metric}
\end{figure}

\begin{figure}[h]
    \centering
    % \vspace{-10pt}
    \includegraphics[width =1.0\linewidth]{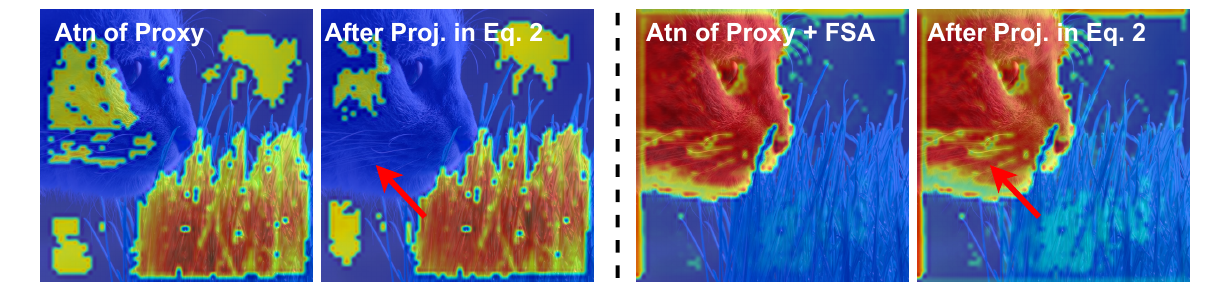}
    \caption{Attention visualization of Fig.2 after Proj. in Eq.2.}
    \label{fig:attn_vis}
\end{figure}

\begin{figure*}[h]
\centering
    %%%%%%%%%%%%%%%%%%%% Row 1 %%%%%%%%%%%%%%%%%%%%
    \begin{subfigure}{0.16\textwidth}
        \includegraphics[width=\linewidth]{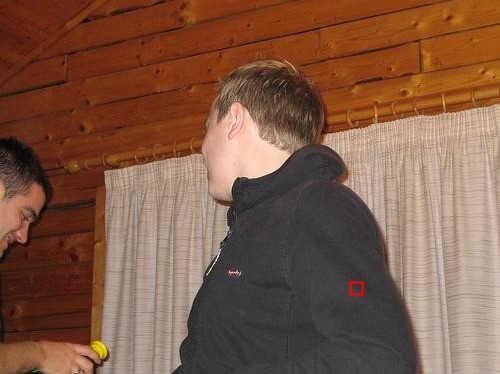}
    \end{subfigure}
    \begin{subfigure}{0.16\textwidth}
        \includegraphics[width=\linewidth]{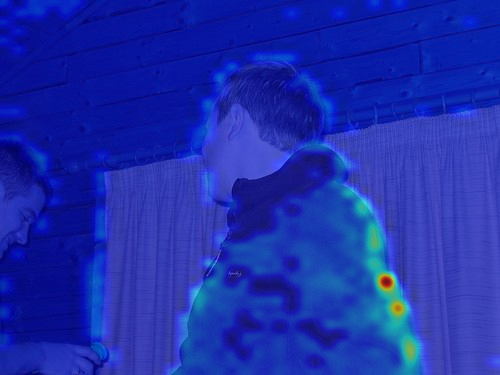}
    \end{subfigure}
    \begin{subfigure}{0.16\textwidth}
        \includegraphics[width=\linewidth]{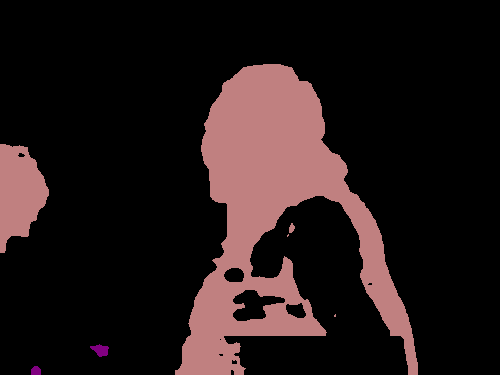}
    \end{subfigure}
    \begin{subfigure}{0.16\textwidth}
        \includegraphics[width=\linewidth]{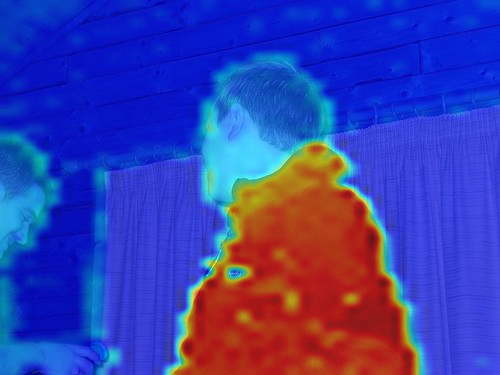}
    \end{subfigure}
    \begin{subfigure}{0.16\textwidth}
        \includegraphics[width=\linewidth]{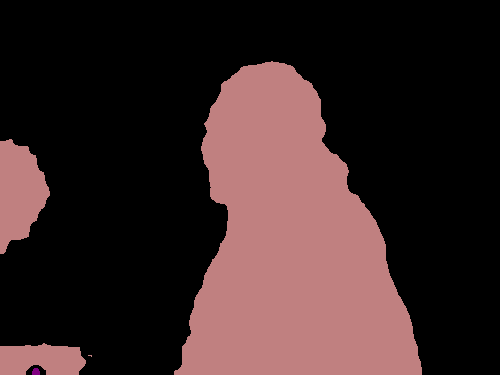}
    \end{subfigure}
    \begin{subfigure}{0.16\textwidth}
        \includegraphics[width=\linewidth]{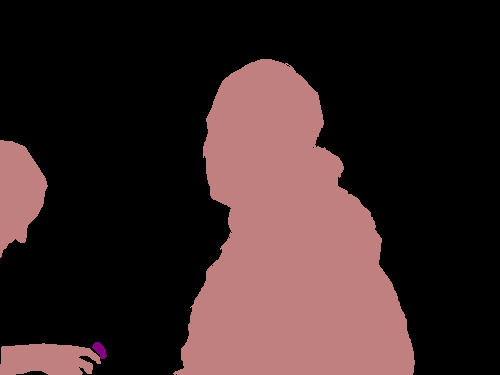}
    \end{subfigure}

    %%%%%%%%%%%%%%%%%%%% Row 2 %%%%%%%%%%%%%%%%%%%%
    \begin{subfigure}{0.16\textwidth}
        \includegraphics[width=\linewidth]{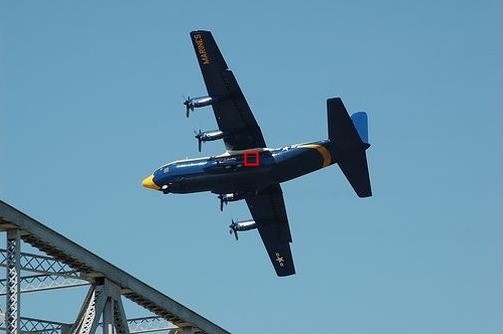}
    \end{subfigure}
    \begin{subfigure}{0.16\textwidth}
        \includegraphics[width=\linewidth]{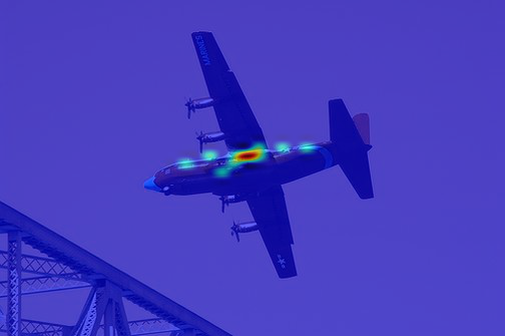}
    \end{subfigure}
    \begin{subfigure}{0.16\textwidth}
        \includegraphics[width=\linewidth]{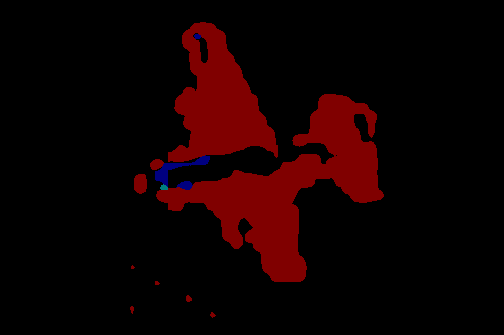}
    \end{subfigure}
    \begin{subfigure}{0.16\textwidth}
        \includegraphics[width=\linewidth]{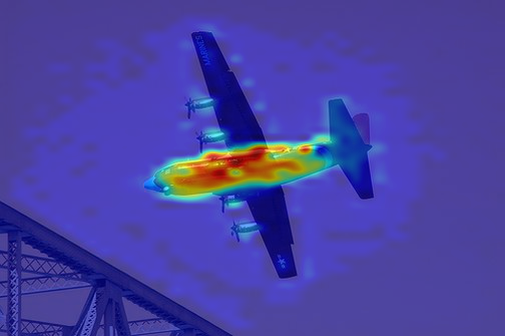}
    \end{subfigure}
    \begin{subfigure}{0.16\textwidth}
        \includegraphics[width=\linewidth]{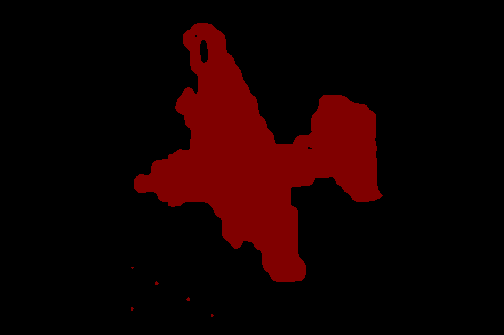}
    \end{subfigure}
    \begin{subfigure}{0.16\textwidth}
        \includegraphics[width=\linewidth]{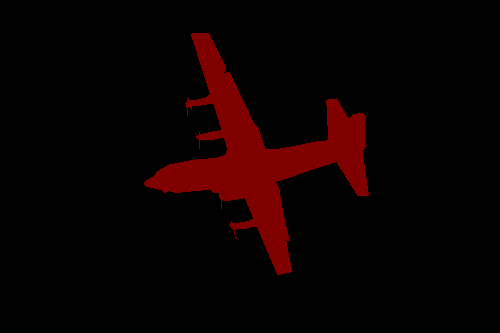}
    \end{subfigure}

    %%%%%%%%%%%%%%%%%%%% Row 3 %%%%%%%%%%%%%%%%%%%%
    \begin{subfigure}{0.16\textwidth}
        \includegraphics[width=\linewidth]{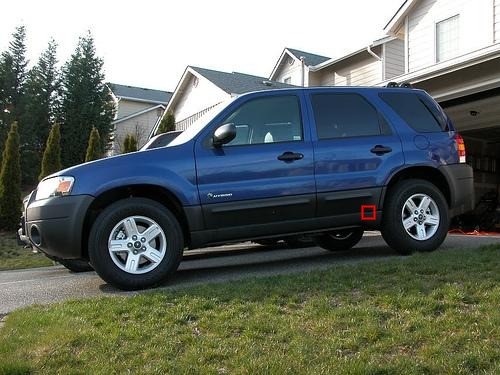}
    \end{subfigure}
    \begin{subfigure}{0.16\textwidth}
        \includegraphics[width=\linewidth]{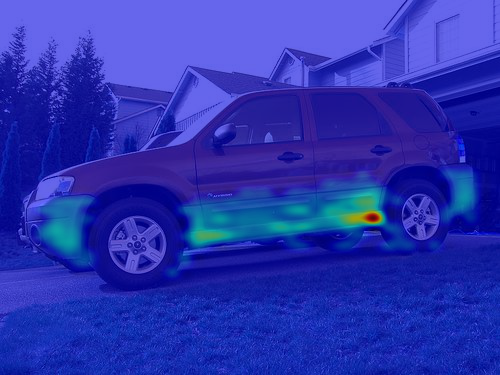}
    \end{subfigure}
    \begin{subfigure}{0.16\textwidth}
        \includegraphics[width=\linewidth]{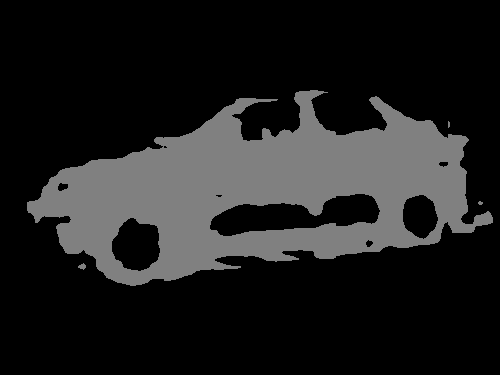}
    \end{subfigure}
    \begin{subfigure}{0.16\textwidth}
        \includegraphics[width=\linewidth]{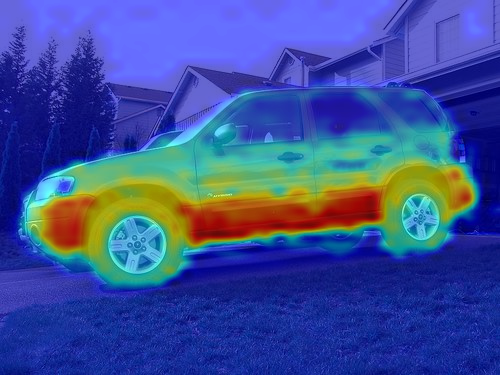}
    \end{subfigure}
    \begin{subfigure}{0.16\textwidth}
        \includegraphics[width=\linewidth]{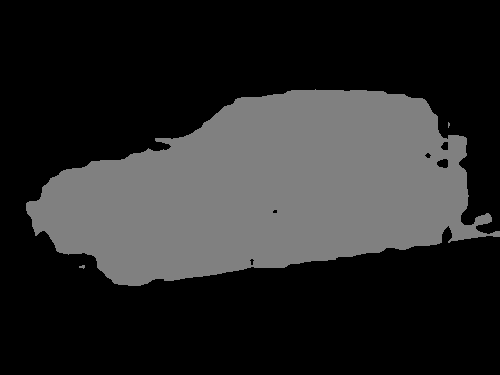}
    \end{subfigure}
    \begin{subfigure}{0.16\textwidth}
        \includegraphics[width=\linewidth]{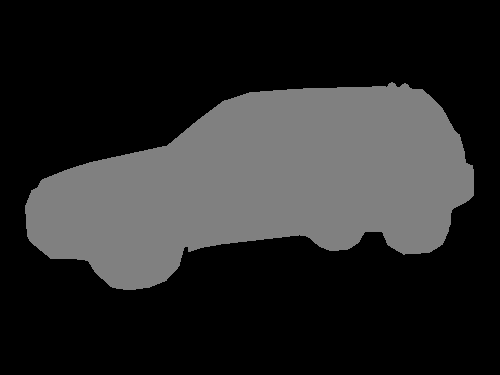}
    \end{subfigure}

    %%%%%%%%%%%%%%%%%%%% Row 4 %%%%%%%%%%%%%%%%%%%%
    \begin{subfigure}{0.16\textwidth}
        \includegraphics[width=\linewidth]{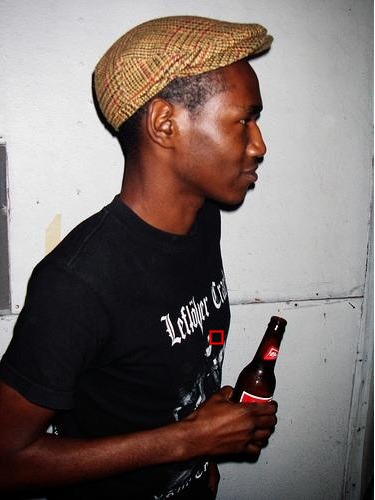}
    \end{subfigure}
    \begin{subfigure}{0.16\textwidth}
        \includegraphics[width=\linewidth]{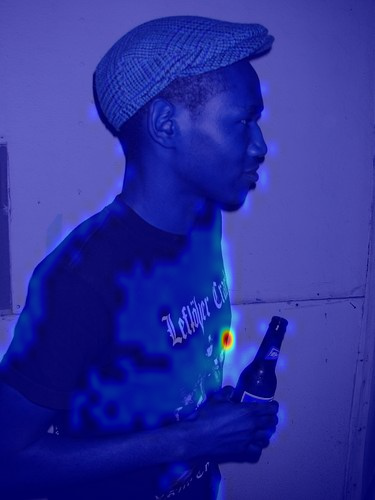}
    \end{subfigure}
    \begin{subfigure}{0.16\textwidth}
        \includegraphics[width=\linewidth]{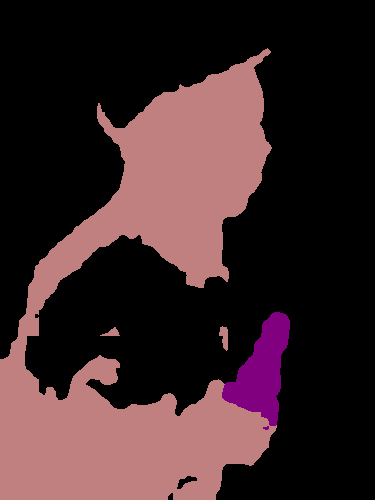}
    \end{subfigure}
    \begin{subfigure}{0.16\textwidth}
        \includegraphics[width=\linewidth]{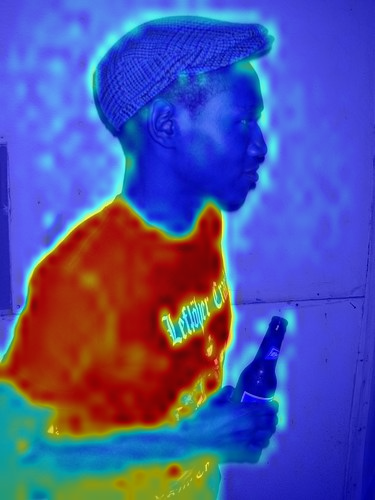}
    \end{subfigure}
    \begin{subfigure}{0.16\textwidth}
        \includegraphics[width=\linewidth]{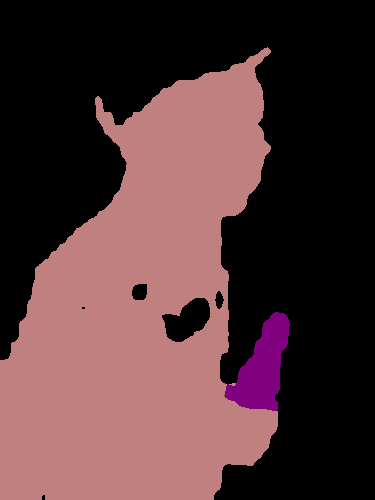}
    \end{subfigure}
    \begin{subfigure}{0.16\textwidth}
        \includegraphics[width=\linewidth]{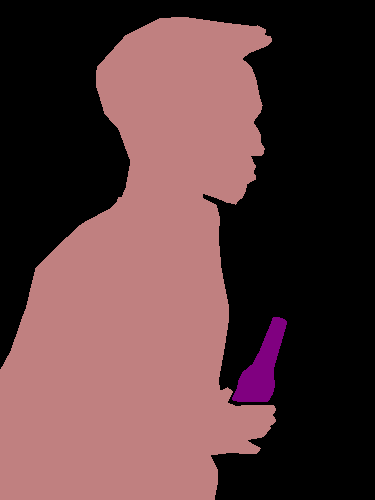}
    \end{subfigure}

    % %%%%%%%%%%%%%%%%%%%% Row 5 %%%%%%%%%%%%%%%%%%%%
    % \begin{subfigure}{0.16\textwidth}
    %     \includegraphics[width=\linewidth]{final_ims/supp_voc21/2007_002234_input_ref.jpg}
    % \end{subfigure}
    % \begin{subfigure}{0.16\textwidth}
    %     \includegraphics[width=\linewidth]{final_ims/supp_voc21/2007_002234_old_1123.png}
    % \end{subfigure}
    % \begin{subfigure}{0.16\textwidth}
    %     \includegraphics[width=\linewidth]{final_ims/supp_voc21/2007_002234_predicted_mask_proxy.png}
    % \end{subfigure}
    % \begin{subfigure}{0.16\textwidth}
    %     \includegraphics[width=\linewidth]{final_ims/supp_voc21/2007_002234_new_1123.png}
    % \end{subfigure}
    % \begin{subfigure}{0.16\textwidth}
    %     \includegraphics[width=\linewidth]{final_ims/supp_voc21/2007_002234_predicted_mask.png}
    % \end{subfigure}
    % \begin{subfigure}{0.16\textwidth}
    %     \includegraphics[width=\linewidth]{final_ims/supp_voc21/2007_002234_gt_mask.png}
    % \end{subfigure}

    %%%%%%%%%%%%%%%%%%%% Row 6 %%%%%%%%%%%%%%%%%%%%
    \begin{subfigure}{0.16\textwidth}
        \includegraphics[width=\linewidth]{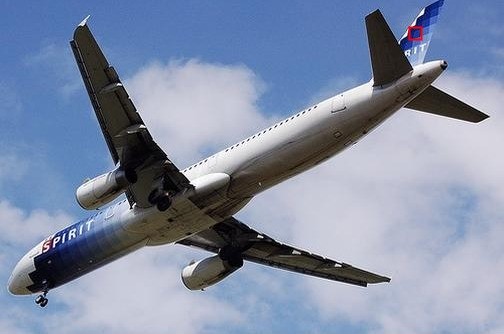}
    \end{subfigure}
    \begin{subfigure}{0.16\textwidth}
        \includegraphics[width=\linewidth]{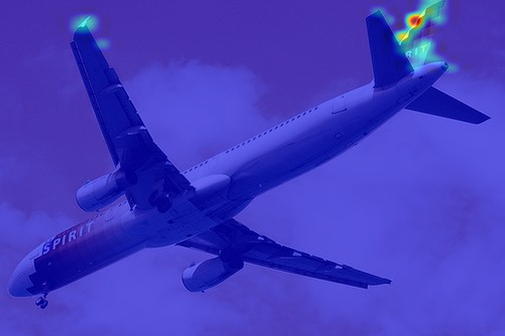}
    \end{subfigure}
    \begin{subfigure}{0.16\textwidth}
        \includegraphics[width=\linewidth]{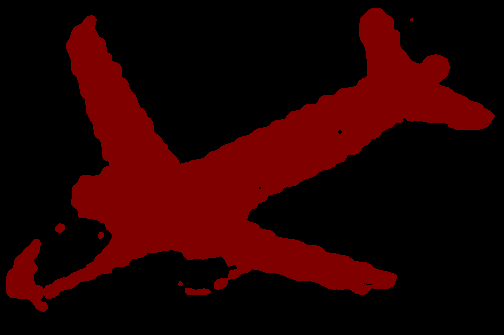}
    \end{subfigure}
    \begin{subfigure}{0.16\textwidth}
        \includegraphics[width=\linewidth]{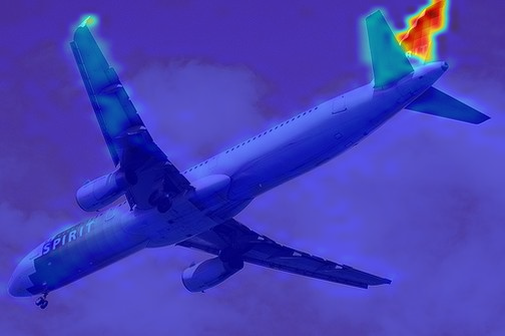}
    \end{subfigure}
    \begin{subfigure}{0.16\textwidth}
        \includegraphics[width=\linewidth]{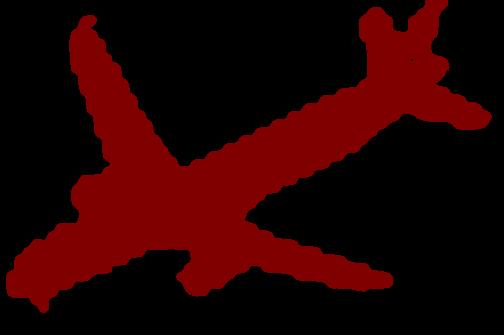}
    \end{subfigure}
    \begin{subfigure}{0.16\textwidth}
        \includegraphics[width=\linewidth]{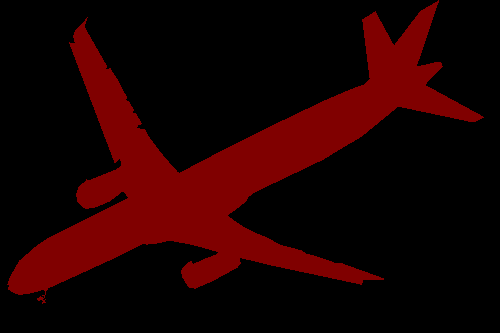}
    \end{subfigure}

    % %%%%%%%%%%%%%%%%%%%% Row 7 %%%%%%%%%%%%%%%%%%%%
    % \begin{subfigure}{0.16\textwidth}
    %     \includegraphics[width=\linewidth]{final_ims/supp_voc21/2007_000121_input_ref.jpg}
    % \end{subfigure}
    % \begin{subfigure}{0.16\textwidth}
    %     \includegraphics[width=\linewidth]{final_ims/supp_voc21/2007_000121_old_710.png}
    % \end{subfigure}
    % \begin{subfigure}{0.16\textwidth}
    %     \includegraphics[width=\linewidth]{final_ims/supp_voc21/2007_000121_predicted_mask_proxy.png}
    % \end{subfigure}
    % \begin{subfigure}{0.16\textwidth}
    %     \includegraphics[width=\linewidth]{final_ims/supp_voc21/2007_000121_new_710.png}
    % \end{subfigure}
    % \begin{subfigure}{0.16\textwidth}
    %     \includegraphics[width=\linewidth]{final_ims/supp_voc21/2007_000121_predicted_mask.png}
    % \end{subfigure}
    % \begin{subfigure}{0.16\textwidth}
    %     \includegraphics[width=\linewidth]{final_ims/supp_voc21/2007_000121_gt_mask.png}
    % \end{subfigure}

    %%%%%%%%%%%%%%%%%%%% Row 8 %%%%%%%%%%%%%%%%%%%%
    \begin{subfigure}{0.16\textwidth}
        \includegraphics[width=\linewidth]{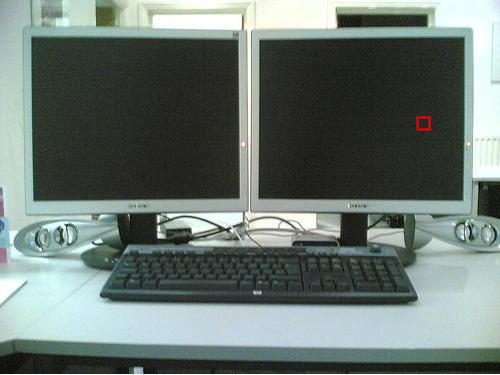}
        \caption*{\makecell{Input \\ image}}
    \end{subfigure}
    \begin{subfigure}{0.16\textwidth}
        \includegraphics[width=\linewidth]{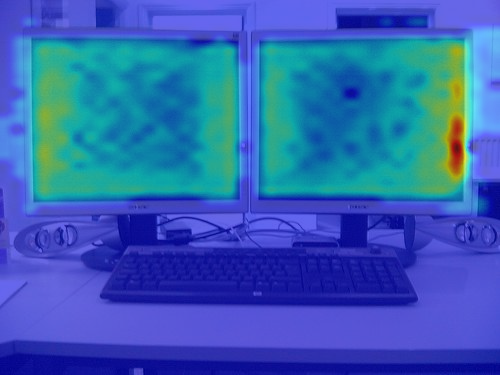}
        \caption*{\makecell{Attention map \\ of ProxyCLIP}}
    \end{subfigure}
    \begin{subfigure}{0.16\textwidth}
        \includegraphics[width=\linewidth]{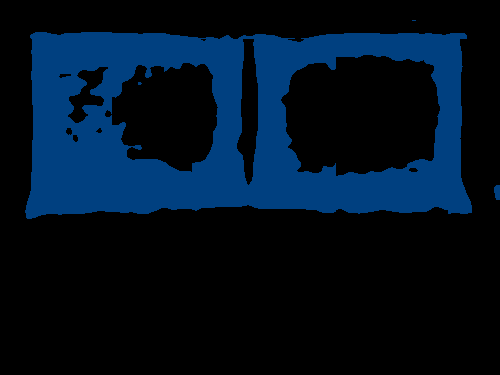}
        \caption*{\makecell{Segmentation map \\ of ProxyCLIP}}
    \end{subfigure}
    \begin{subfigure}{0.16\textwidth}
        \includegraphics[width=\linewidth]{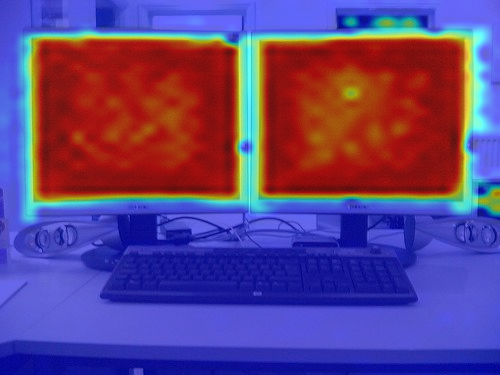}
         \caption*{\makecell{Our self-adapted \\ attention map}}
    \end{subfigure}
    \begin{subfigure}{0.16\textwidth}
        \includegraphics[width=\linewidth]{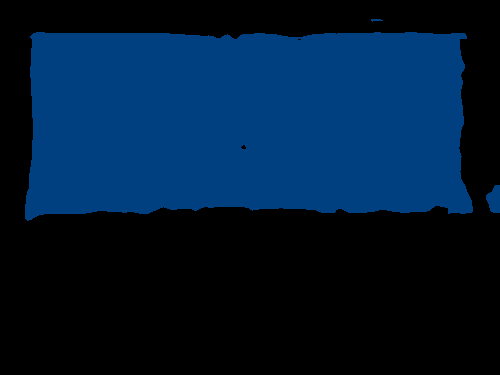}
        \caption*{\makecell{ProxyCLIP \\ + our FSA}}
    \end{subfigure}
    \begin{subfigure}{0.16\textwidth}
        \includegraphics[width=\linewidth]{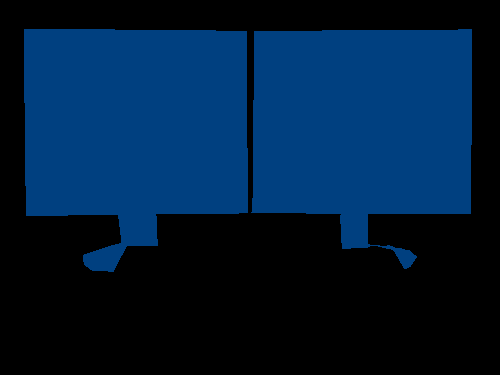}
        \caption*{\makecell{Ground truth \\ segmentation map}}
    \end{subfigure}
    
\caption{\textbf{Comparative visualization of segmentation results: ProxyCLIP vs. integration with our FSA.} The attention maps (second and fourth columns) correspond to the reference patch shown in the first column. ProxyCLIP produces segmentation maps with holes within the same object due to weak attention across regions of the object. In contrast, integrating our FSA effectively aggregates similar patches, enabling the correction of missegmented regions.
}
\label{fig:supp_voc}
\end{figure*}

In the main paper, we have illustrated the semantic coherence retention. To quantify subsequent degradation, we introduce a new metric: using $Attn^{init}$ as reference, for each patch $i$, we get its most attended patch $j$. After each operation in Eq.2 (residuals, FFNs), we compute pairwise token similarities and check whether $j$ remains among the top-$10$ similar patches to $i$. Fig.~\ref{fig:new_metric} illustrates this process and the metric drops (ave of 8 datasets) sharply after residuals in MaskCLIP, indicating noise injection~\textcolor{NavyBlue}{[21]}. In contrast, our FSA better preserves spatial coherence. Fig.~\ref{fig:attn_vis} compares attention maps (Fig.2) after the proj: although both methods reduce focus on the cat’s face, our improved intermediate attention provides greater \textit{resistance to degradation.}

\section{Similarity metric}
Table~\ref{tab:ablation_similarity} compares cosine similarity and KL divergence for computing logit similarity. KL divergence proves more effective due to its ability to assess full distributions and highlight differences in probabilistic outputs, making it better suited for capturing detailed semantic coherence and enabling effective feedback adaptation.

\begin{table}[t]
    \centering
    \begin{tabular}{c|ccc}
        \toprule
         Similarity metric & ViT-B/16 & ViT-L/14 & ViT-H/14 \\
        \midrule
        Cosine & 43.2 & 43.3 & 45.4 \\
        KL divergence & 43.3 & 43.6 & 45.8\\
        \bottomrule
    \end{tabular}
    \caption{\textbf{Sensitivity on similarity metric.} KL divergence evaluates entire distributions and emphasizes differences in probabilistic outputs, making it ideal for capturing detailed semantic coherence and supporting effective feedback adaptation.
}
    \label{tab:ablation_similarity}
\end{table}

\section{Impact of different configuration of CLIP}

ProxyCLIP and ClearCLIP omit residual and FFN modules, identified as sources of noisy segmentation~\cite{lan2024proxyclip, lan2024clearclip}, thus better preserving spatial consistency than MaskCLIP or SCLIP, which retain them. As our method primarily enhances semantic consistency, it yields larger improvements on baselines with weaker spatial coherence. As shown in Tab.~\ref{tab:general}, FSA improves MaskCLIP under different configurations of CLIP, though the margin is smaller in the latter.

\begin{table}[h]

\centering
\vspace{-10pt}
    % \tiny
    \small
    \tabcolsep3pt
    \begin{tabular}{c|lll}
        \toprule
        Methods & ViT-B/16 & ViT-L/14 & ViT-H/14 \\
        \midrule
        MaskCLIP & 27.9 & 13.9 & 19.3 \\
        \cellcolor{gray!20.0}{+FSA} & \cellcolor{gray!20.0}{\bestimprovered{35.8}{7.9}} & \cellcolor{gray!20.0}{\bestimprovered{32.6}{18.7}} & \cellcolor{gray!20.0}{\bestimprovered{33.4}{14.1}} \\
        \hline
        MaskCLIP (w/o FFN, Res) & 29.5 & 29.7 & 29.8 \\
        \cellcolor{gray!20.0}{+FSA} & \cellcolor{gray!20.0}{\bestimprovered{36.8}{7.3}} & \cellcolor{gray!20.0}{\bestimprovered{34.1}{4.4}} & \cellcolor{gray!20.0}{\bestimprovered{34.7}{4.9}}\\
        \bottomrule
    \end{tabular}
    \caption{\textbf{Improvement over MaskCLIP with different configurations.} Average mIoU reported on 8 datasets.} 
    \label{tab:general}
% \vspace{-14pt}
\end{table}

\section{Additional speed analysis}
Following Clear/Mask/SCLIP, we modify only the \textit{last layer}, incurring a 4.3--11.7\% overhead depending on layer count (Tab.~\ref{tab:cost_supp}).

\sethlcolor{gray!20}
\begin{table}[H]
    \centering
    % \tiny
    \small
    \tabcolsep0.9pt
    \begin{tabular}{c|ccc}
        \toprule
        Methods & B/16(12-layers) & L/14(24-layers) & H/14(32-layers) \\
        \midrule
        Clear/\hl{+FSA} & 4.9/\hl{5.4}	& 13.1/\hl{13.9} & 21.1/\hl{22.2} \\
        Mask/\hl{+FSA} & 5.1/\hl{5.7} & 13.4/\hl{14.1} & 21.9/\hl{22.9} \\
        SCLIP/\hl{+FSA} & 5.2/\hl{5.7} & 13.4/\hl{14.2} & 22.0/\hl{23.0} \\
        \bottomrule
    \end{tabular}
    \caption{Speed (ms) on V100 GPU with 224x224 input.} 
    \label{tab:cost_supp}
\end{table}

% \clearpage
% \newpage

% {
%     \small
%     \bibliographystyle{ieeenat_fullname}
%     \bibliography{main}
% }

\end{document}